\documentclass[10pt,twocolumn,letterpaper]{article}

\usepackage{cvpr}              

\usepackage{amsthm}

\usepackage{bm}
\usepackage{algorithm}
\usepackage{algpseudocode}
\usepackage{algorithmicx}
\definecolor{cvprblue}{rgb}{0.21,0.49,0.74}
\usepackage[pagebackref,breaklinks,colorlinks,allcolors=cvprblue]{hyperref}
\usepackage{adjustbox}
\usepackage{colortbl}
\def\paperID{} 
\def\confName{CVPR}
\def\confYear{2026}

\title{HTTM: Head-wise Temporal Token Merging for Faster VGGT}

\author{
Weitian Wang$^{1,2}$\thanks{Emails: \{weitian.wang, lukas.meiner, shubham.rai, cecilia.delaparra\}@bosch.com, akash.kumar@ruhr-uni-bochum.de},
Lukas Meiner$^{1}$, Rai Shubham$^1$, Cecilia De La Parra$^1$, Akash Kumar$^2$\\
$^1$Robert Bosch GmbH, Germany,
$^2$Ruhr University Bochum, Germany
}

\begin{document}
\maketitle
\begin{abstract}
The Visual Geometry Grounded Transformer (VGGT) marks a significant leap forward in 3D scene reconstruction, as it is the first model that directly infers all key 3D attributes (camera poses, depths, and dense geometry) jointly in one pass. However, this joint inference mechanism requires global attention layers that perform all-to-all attention computation on tokens from all views. For reconstruction of large scenes with long-sequence inputs, this causes a significant latency bottleneck. In this paper, we propose head-wise temporal merging (HTTM), a training-free 3D token merging method for accelerating VGGT.
Existing merging techniques merge tokens uniformly across different attention heads, resulting in identical tokens in the layers' output, which hinders the model's representational ability. HTTM tackles this problem by merging tokens in multi-head granularity, which preserves the uniqueness of feature tokens after head concatenation. Additionally, this enables HTTM to leverage the spatial locality and temporal correspondence observed at the head level to achieve higher merging ratios with lower merging costs compared to existing methods. Thus, HTTM achieves up to $7\times$ acceleration over the original VGGT with negligible performance drops in a GPU-based inference.

\vspace{-1em}
\end{abstract}
    
\section{Introduction}

\begin{figure}[htbp]
  \centering
    \centering
    \begin{subfigure}[t]{0.56\linewidth}  
      \includegraphics[width=\linewidth]{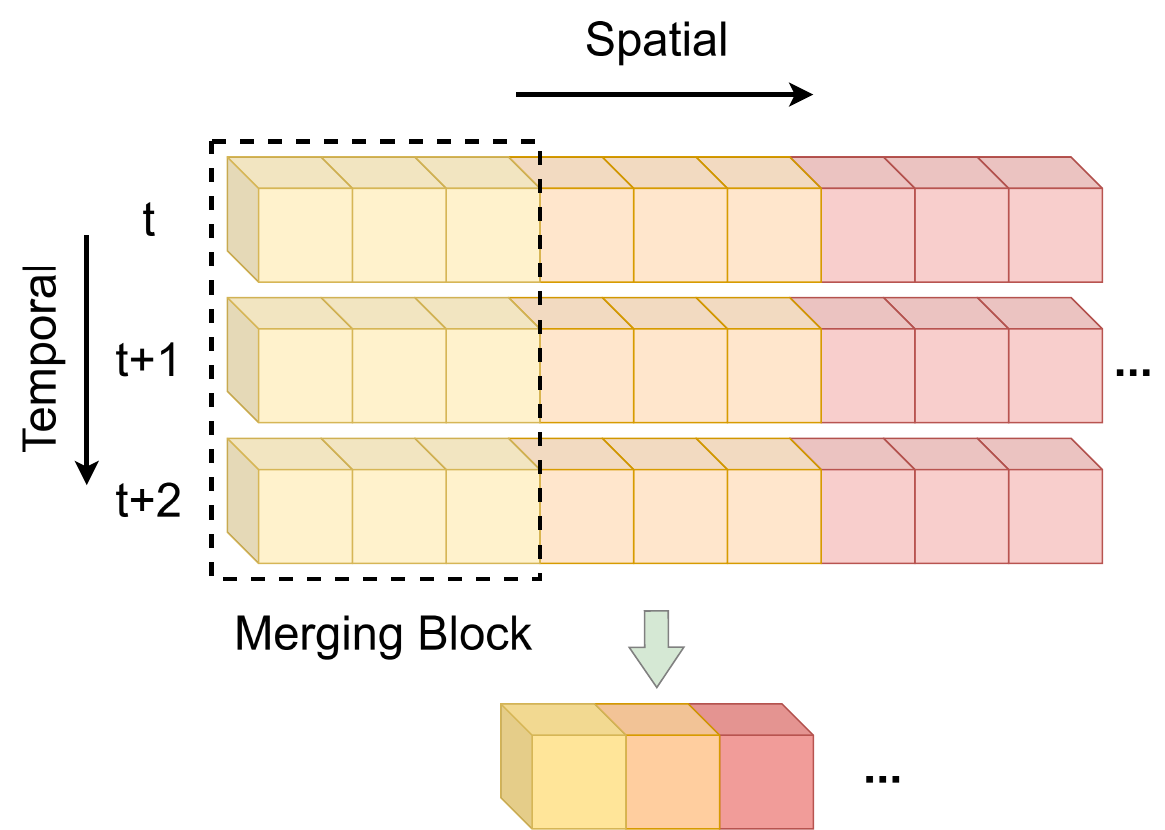}
      \label{fig:wo_reordering}
    \end{subfigure}
    \hfill
    \begin{subfigure}[t]{0.43\linewidth}  
      \includegraphics[width=\linewidth]{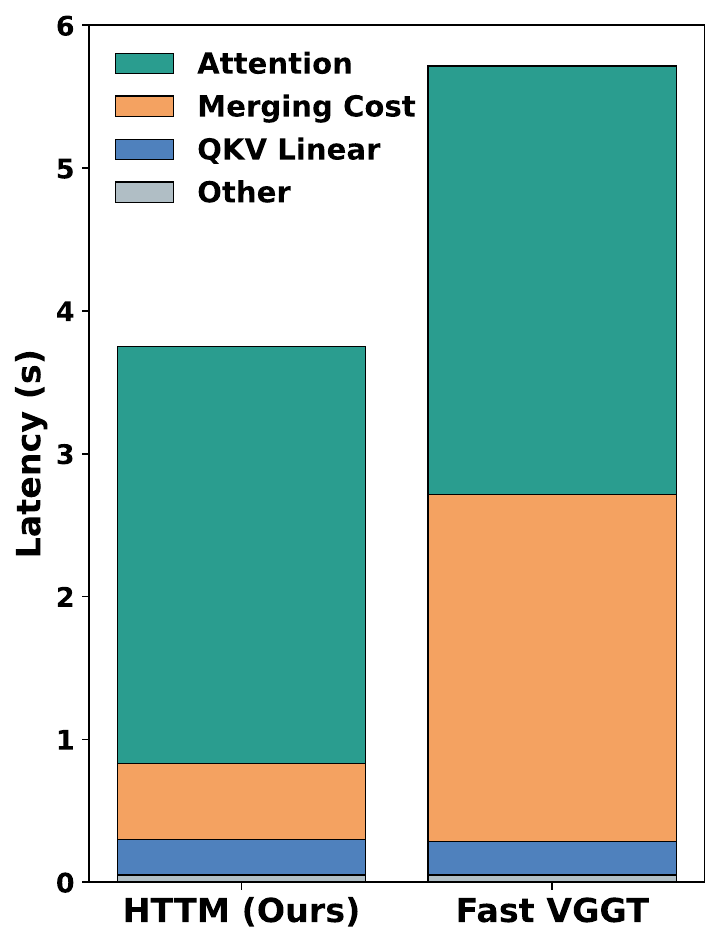}
      \label{fig:cost_components}
    \end{subfigure}
    \caption{HTTM forms spatio-temporal merging blocks that jointly consider neighboring tokens across consecutive frames. This design exploits temporal coherence and spatial redundancy to merge tokens efficiently. With the same merging ratio, HTTM reduces the merging cost by \textbf{4.58$\times$}.}
    \label{fig:intro_fig}
\vspace{-1em}
\end{figure}

The Visual Geometry Grounded Transformer (VGGT)~\cite{wang2025vggt} is a recently proposed feed-forward transformer model that directly infers all key 3D attributes of a scene from a variable number of views. By this direct inference, VGGT can outperform state-of-the-art methods while avoiding costly visual-geometry post-processing methods, marking an important breakthrough in 3D computer vision.

One of the key designs of VGGT is its alternating frame-wise and global attention. In the global attention layers, all tokens from different views participate in the multi-head attention computation. In practice, this approach results in extremely long token sequences (more than 20k tokens) even for small scenes. Thus, the global attention layers become the main latency bottleneck of VGGT, limiting its efficiency in medium and large scene reconstruction.

Motivated by the development of long-context large language models (LLMs) and vision language models (VLMs), many methods~\cite{xiao2024efficient, child2019generating, zaheer2020big, Kitaev2020Reformer, vyas2020fast} have been proposed to ease the high computational costs of long-sequence attention layers. These methods are mainly sparsity-based approaches, aiming to exploit that attention scores in LLMs and VLMs tend to concentrate on a small set of tokens. However, the sparsity level of VGGT's global attention layers is lower compared to the attention layer in LLM, as shown in Fig.~\ref{fig:attn_dist}, which limits the latency improvement of these methods when applied to VGGT.


On the other hand, the narrow attention distribution patterns that appear in VGGT favor similarity-based methods, as the attention weight distributions do not change drastically from token to token. ToMe~\cite{bolya2022token} gave rise to a variety of token merging approaches that accelerate transformers by merging redundant tokens based on feature similarity. Following this work, a number of extensions~\cite{kim2024tofu, chen2023diffrate, bolya2023tomesd, wang2025dymu, shen2025fastvggt} have been proposed, demonstrating that similarity-based token reduction can effectively improve efficiency without retraining.


While several approaches~\cite{bolya2023tomesd, shen2025fastvggt} are designed for, or can be adapted to VGGT, they fail to recognize the special head-level similarity pattern of VGGT, and exhibit high merging overhead on long input sequences.

To this end, we propose HTTM (Head-wise Temporal Token Merging), a training-free token merging approach tailored for VGGT's Global Attention layers. HTTM addresses the limitations of existing methods through three key innovations: (1) head-wise merging that allows each attention head to merge tokens independently, preserving head-specific information and avoiding feature collapse after concatenation; (2) block-wise token merging that reduces the token matching cost compared to global matching strategies. (3) temporal reordering that reorganizes tokens into spatio-temporal blocks to exploit both spatial redundancy and temporal coherence, improving merging quality in fixed merging block size; (4) head-wise adaptive outlier filtering that filters outliers across all heads under a global budget.
These designs enable HTTM to achieve notable speedups while maintaining high reconstruction quality.

Our main contributions are summarized as follows:
\begin{itemize}

\item We conduct systematic explorations of token merging in VGGT, revealing distinct similarity patterns along both spatial and temporal dimensions. Through extensive analysis, we show that temporal correlations exhibit higher redundancy, motivating the need for temporally aware merging strategies.

\item Recognizing the main computational overhead in existing token merging methods, we analyze the trade-off between merging cost and merging quality, and propose a block-wise merging strategy with temporal reordering that largely reduces merging cost while preserving merging quality by aligning spatially and temporally correlated tokens.


\item We introduce an adaptive outlier filtering mechanism that filters outliers across all attention heads under a global budget, allocating more budget to heads with higher outlier density and improving overall quality with minimal overhead.

\end{itemize}

\section{Related Work}
\subsection{Efficient Attention for Long Sequences}
\begin{figure}
    \centering
    \includegraphics[width=0.6\linewidth]{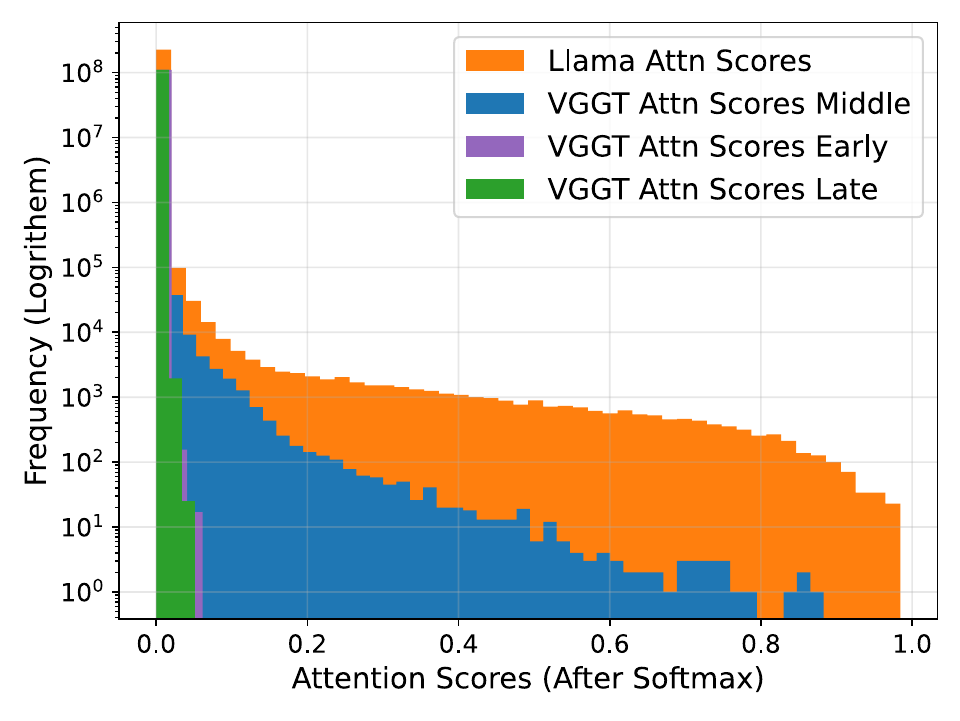}
     \caption{Attention score distribution comparison between VGGT and Llama 3.1 8B\cite{grattafiori2024llama3herdmodels}. 
     The distribution of attention scores in VGGT is heavily concentrated around low values in both its early and late layers. In the middle layers, attention distribution is still more skewed towards lower values compared to Llama.
     }
    \label{fig:attn_dist}
    \vspace{-0.5cm}
\end{figure}

To address the quadratic complexity of attention layers in transform-based models over long sequences, sparse attention methods have recently gained popularity. Utilizing the inherently sparse attention mechanism in LLMs, methods like Sparse Transformer~\cite{child2019generating} and BigBird~\cite{zaheer2020big} accelerate inference and extend maximum sequence lengths through local or block-based attention but often require retraining. StreamingLLM~\cite{liu2023ring} enables LLMs to handle unlimited texts without fine-tuning by retaining the attention sinks (initial tokens) and the recent tokens. SparseVLM~\cite{zhang2024sparsevlm} accelerates VLMs through a text-guided training-free token pruning mechanism. For vision transformers, SparseViT~\cite{chen2023sparsevit} and MixA-Q~\cite{wang2025mixa} discard or compress unimportant tokens with small magnitudes for latency improvements. However, these methods are all dependent on the sparse attention pattern, which VGGT doesn't exhibit distinctly as shown in Fig.~\ref{fig:attn_dist}.

    

\subsection{Accelerating Visual Geometry Models}
    Feed-forward models for 3D reconstruction tasks like DUSt3R~\cite{wang2024dust3r}, MASt3R~\cite{leroy2024mast3r}, and VGGT~\cite{wang2025vggt} have emerged as strong alternatives to traditional multi-view reconstruction pipelines. One of the key designs of VGGT is its alternating global attention and frame attention layer. In frame attention layers, tokens interact within each frame independently. In global attention layers, tokens interact globally across all frames, helping the model build global 3D correspondence. However, as input sequences grow longer, VGGT's global attention becomes a computational bottleneck~\cite{shen2025fastvggt, wang2025fastervggtblocksparse}.
    
    Online methods like CUST3R~\cite{wang2025continuous} and TTT3R~\cite{chen2025ttt3r} maintain a global state that is updated for each new input frame, thus avoiding the all-to-all attention. Nevertheless, as reported in TTT3R~\citet{chen2025ttt3r}, their performance is still not comparable to offline methods like VGGT. 

    While a few approaches~\cite{shen2025fastvggt, wang2025fastervggtblocksparse} try to accelerate VGGT by applying established methods for ViT or LLM token sequence compression, their analysis of VGGT's specific redundancies and resulting opportunities for compression is limited.
    FastVGGT~\cite{shen2025fastvggt} proposes to use the token merging from ToMeSD~\cite{bolya2023tomesd} to accelerate the global attention layers. Although it achieves high latency improvements, FastVGGT fails to adapt the merging method from the 2D vision tasks to the similarity patterns of VGGT, limiting its performance. 
    Block-sparse global attention \cite{wang2025fastervggtblocksparse} exploits the attention sparsity found in VGGT's middle layers. However, as shown in Fig~\ref{fig:attn_dist}, the sparsity level of VGGT is lower compared to the LLMs, limiting the latency improvements it can achieve without hurting the performance.

\subsection{Token Merging Methods}
    
    ToMe~\cite{bolya2022token} pioneered token merging to reduce the number of tokens for ViTs, employing bipartite soft matching of tokens based on key similarity averaged across attention heads. Extensions such as ToMeSD~\cite{bolya2023tomesd} enable token merging for diffusion models by introducing unmerging operations to restore dense token sequences, while ToFu~\cite{kim2024tofu} allows for both pruning and merging of tokens.
    Token merging has also been extended to video domains~\cite{lee2024videotokenmerging, feng2024efficientvideotransformers, hyun2025multi}. These approaches explore the use of spatio-temporal merging, leveraging the redundancy across video frames to achieve better merging strategies.
    HTTM shares the same underlying intuition with these methods. However, given that token merging and unmerging need to be applied to every layer in VGGT, our method specifically focuses on limiting the cost of token similarity computations instead of devising sophisticated strategies for better merging quality.
    


\section{Head-wise Temporal Token Merging}
\subsection{Observations and Insights}
\label{sec:observationsandinsights}

Before going into details of our head-wise temporal token merging method, we first present the observations and intuitions that led us to this particular approach. Unless otherwise stated, all the similarity matrices or similarity patterns shown in this section are self-cosine-similarity matrices over a sequence of tokens.


\begin{figure}[hbtp]
    \centering
    \includegraphics[width=0.7\linewidth]{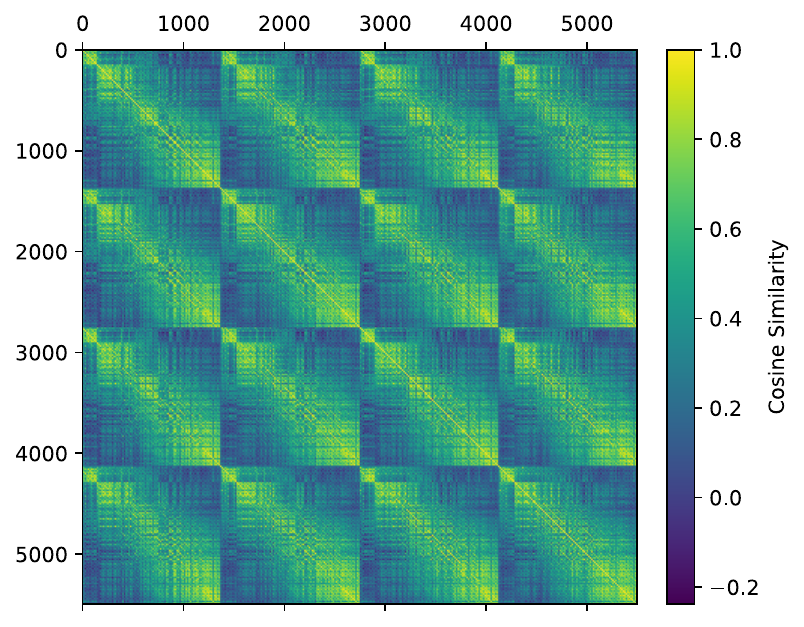}
    \caption{Cosine similarity patterns averaged across all heads between query tokens of 4 adjacent frames. High similarities observed along the block diagonals indicate that tokens within the same spatial region (local areas) and corresponding locations across consecutive frames share highly similar features. 
    }
    \label{fig:token_sim}
    \vspace{-0.6cm}
\end{figure}

\paragraph{Spatial \& Temporal Token Similarity}  
Fig. \ref{fig:token_sim} illustrates the cosine similarity patterns between all the query tokens in the first 4 frames of a scene from the NRGBD~\cite{azinovic2022neural} dataset. It can be observed that high similarity scores concentrate near the main and off-diagonals. The high similarity scores near the main diagonal indicate that each token mainly attends to similar tokens in its local neighborhood. On the other hand, the high similarity scores near the off-diagonals have offsets of frame length, meaning that the same area of different frames also shows high similarity. We will explore the reason behind these patterns in the following paragraphs.

\paragraph{RoPE Effect}

The strong periodic patterns in Fig.~\ref{fig:token_sim} emerge from the way the Rotary Position Embedding (RoPE) \cite{su2024roformer} is applied at each attention layer of VGGT~\cite{wang2025vggt}. Unlike models such as BERT~\cite{devlin2019bert} or Stable Diffusion~\cite{rombach2022high}, which use a fixed positional embedding added only once at the input, VGGT reapplies RoPE at every layer, which strengthens positional encoding effects throughout the network. In the global attention layers, where tokens from all views interact, RoPE differentiates between individual frames, enhancing spatial distinctiveness and reducing similarity between distant locations. In the frame attention layers, RoPE is applied identically to each frame, so corresponding regions across frames share similar positional encodings, inducing temporal coherence between the same spatial areas of adjacent frames. More discussion on this can be found in the Appendix.

\begin{figure}
    \centering
    \includegraphics[width=0.75\linewidth]{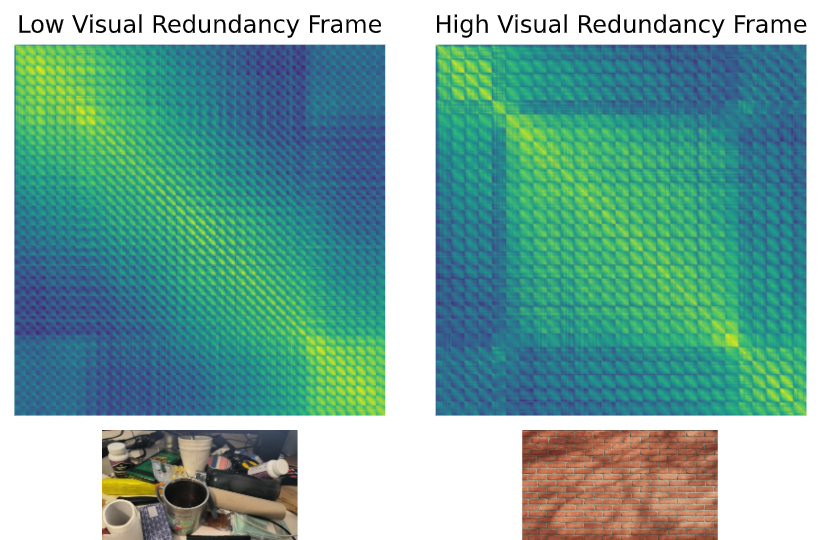}
    \caption{Cosine similarity between query tokens at the 14th deep global attention layer in single-frame reconstruction. The high visual redundancy frame (a wall) shows stronger spatial similarity compared to the low visual redundancy frame (cluttered objects).}
    \label{fig:singleframe}
\end{figure}

\paragraph{Input Similarity Effect}
Although RoPE induces a periodic similarity pattern, intra- and inter-frame similarities also contribute to this observation. 
As shown in Fig.~\ref{fig:singleframe}, we perform single-frame reconstruction on two images with different levels of visual redundancy in pixel patches. At a deep Global Attention layer (14th), the query tokens from the high-redundancy input frame (a wall) show much stronger spatial similarity than the low-redundancy input frame with cluttered objects.

\begin{figure}
    \centering
    \includegraphics[width=\linewidth]{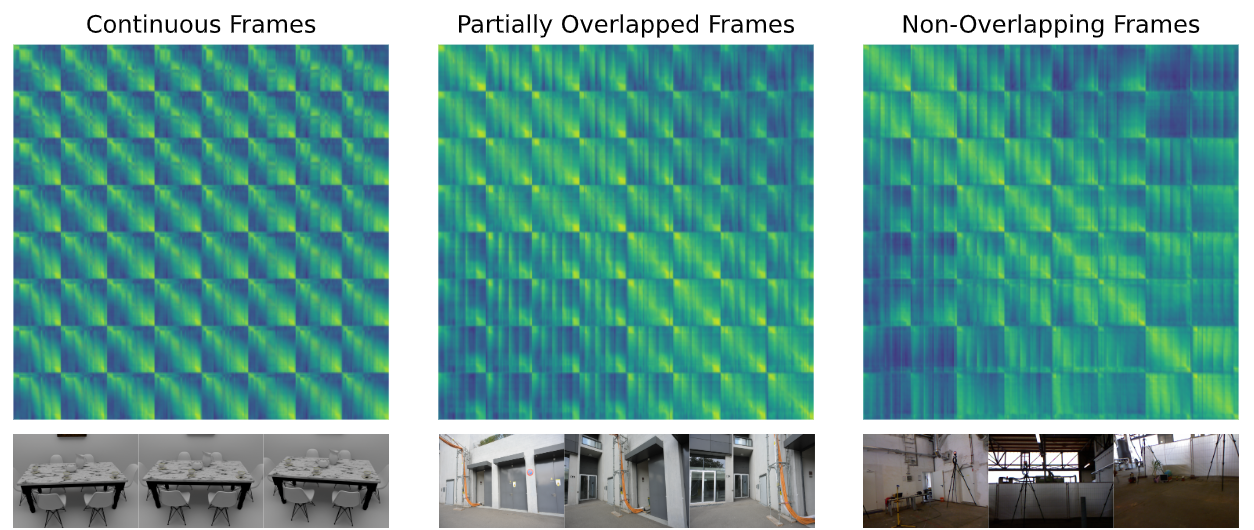}
    \caption{Cosine similarity between query tokens of 8 adjacent frames at the 14th global attention layer. High similarity between input frames leads to high temporal similarity of query tokens, as shown by the high scores on off-diagonals.}
    \label{fig:frames_n_sim}
\end{figure}

In Fig.~\ref{fig:frames_n_sim}, we visualize the query tokens' similarity of a deep layer (14th) across 8 consecutive frames from three 30-frame reconstructions under varying degrees of visual continuity. When the input frames are temporally continuous and highly similar, strong off-diagonal responses appear, indicating that tokens corresponding to the same spatial regions in adjacent frames exhibit high similarity. As the overlap decreases, these off-diagonal structures become weaker and more dispersed, reflecting diminished temporal correspondence between tokens.

\paragraph{Summary}

These findings suggest that the observed similarity pattern originates from two intertwined factors: the architectural effect of RoPE, which enforces spatial distinctness in global attention layers and temporal correspondence in frame attention layers, and the input-level similarity, which propagates through the network and reinforces correlations between spatially corresponding regions. Together, they shape the distinctive spatio–temporal similarity structure of tokens in VGGT.
Hence, we devise a merging strategy that can jointly consider the spatial locality and temporal correspondence.

\begin{figure*}[t]
  \centering
  \includegraphics[width=\linewidth]{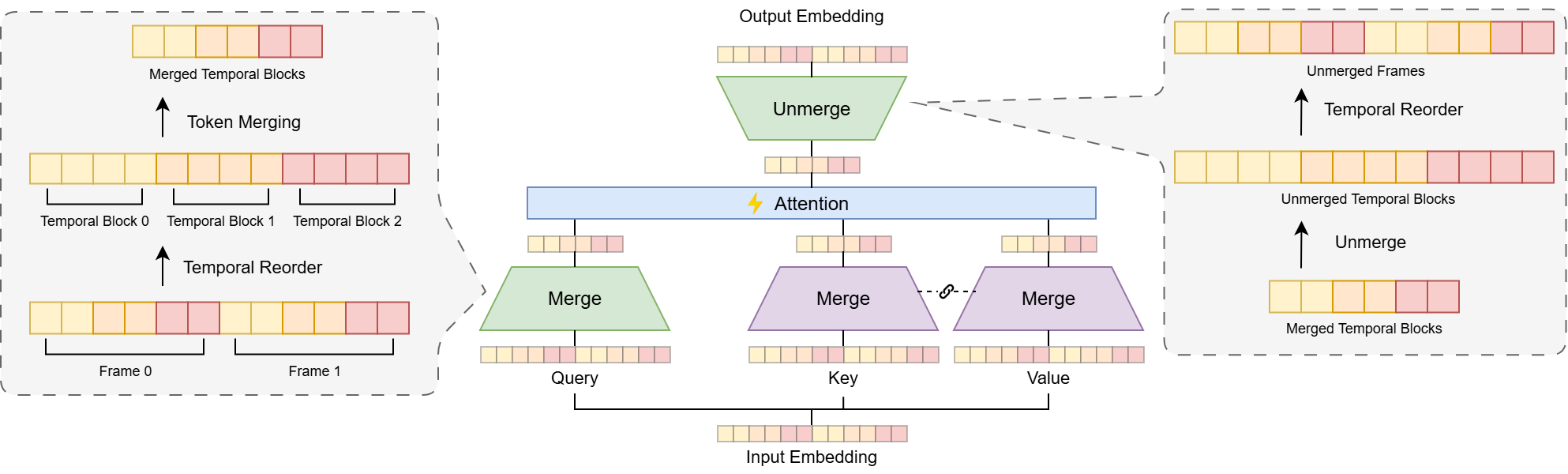}
  \caption{Overview of how HTTM accelerates attention layers by merging QKV tokens. HTTM merges and unmerges Q/K/V tokens before and after entering the attention kernel. Using temporal reordering~\ref{sec:merge}, HTTM forms temporal blocks that consist of similar tokens (denoted with colors) and performs merging and unmerging within these blocks.}
  \label{fig:overview}
\end{figure*}

\subsection{Head-wise Token Merging}
\label{sec:tomesd}
\begin{figure}
    \centering
    \includegraphics[width=0.9\linewidth]{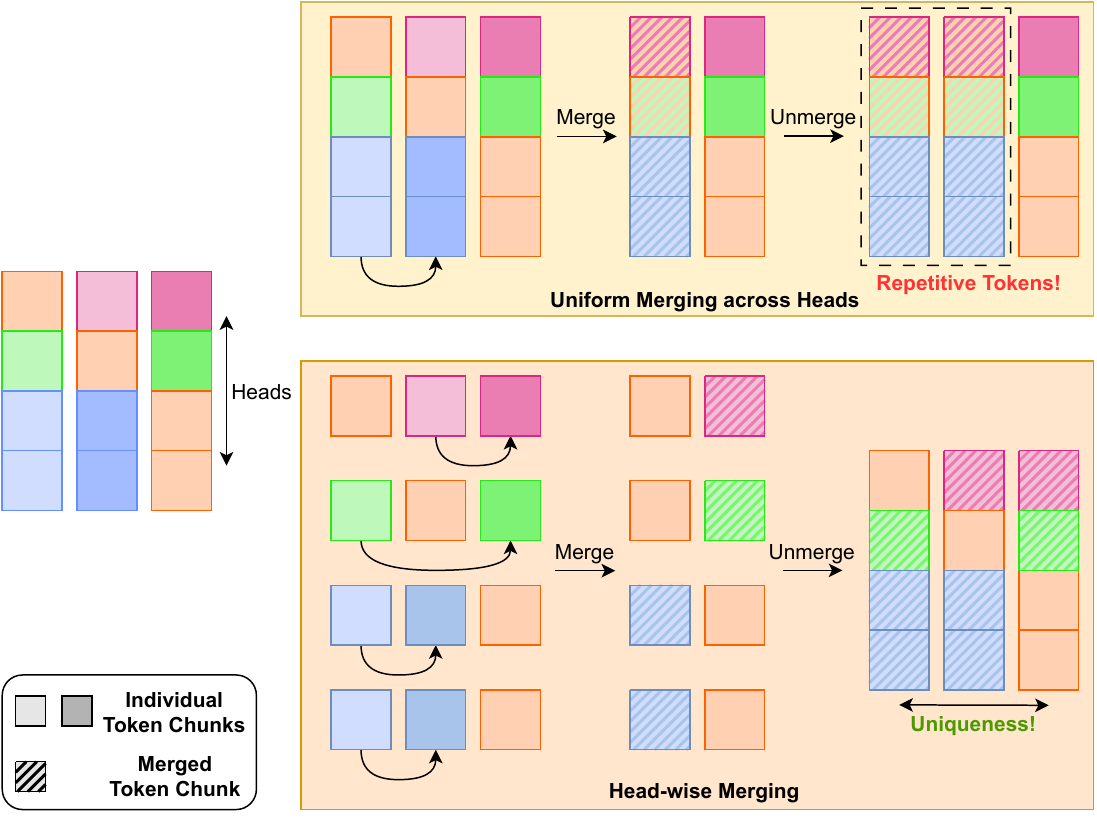}
    \caption{Head-wise merging can better keep the uniqueness of the output embedding. Different shades of the same color represent similar token chunks.}
    \label{fig:repetitive_token}
\end{figure}

Existing token merging methods~\cite{bolya2022token,bolya2023tomesd,shen2025fastvggt} typically adopt a uniform merging strategy for all the heads.
While this strategy simplifies the merging process, 
shared merge topology across heads constrains head-specific aggregation patterns and can reduce representational diversity after head concatenation, as illustrated in Fig.~\ref{fig:repetitive_token}. 
To address this issue, we use a head-wise token merging strategy in HTTM, in which each head performs merging independently according to its own similarity patterns. This head-specific merging enables different heads to combine tokens in distinct ways, producing more diverse and complementary representations after concatenation.

However, the head-wise token merging introduces an overhead that increases proportionally to the number of heads, which is the reason why existing methods avoid this strategy. In Sec.~\ref{sec:merge}, we present our approach to lower the merging overhead by using a temporal merging block.

We now describe the head-wise token merging procedure in HTTM. As shown in Fig.~\ref{fig:overview}, for each head of a multi-head attention layer, its query and key tokens are merged independently using a merge module. Value tokens are merged following the key tokens, as the attention computation requires key-value consistency.

Formally, given a sequence of $N$ input tokens $\mathbf{X} \in \mathbb{R}^{N \times d}$ with embedding dimension $d$, we apply standard multi-head attention projections to obtain 
$\mathbf{Q}, \mathbf{K}, \mathbf{V} \in \mathbb{R}^{h \times N \times d_{\text{head}}}$,
where $h$ is the number of attention heads and $d_{\text{head}} = d / h$ is the embedding dimension per head. For each attention head $i \in \{1, \ldots, h\}$, we then denote its head embeddings $\mathbf{Q}^{(i)}, \mathbf{K}^{(i)}, \mathbf{V}^{(i)} \in \mathbb{R}^{N \times d_{\text{head}}}$. 

\paragraph{Merging}

Our head-wise temporal merging can be understood as a set of functions
$\mathcal{M}_i: \mathbb{R}^{N \times d_{\text{head}}} \to \mathbb{R}^{M \times d_{\text{head}}}$ that reduce the head-wise token sequence length from $N$ to $M$, independently for each head. Following ToMeSD~\cite{bolya2023tomesd}, the tokens are partitioned into disjoint sets of source (\texttt{src}) tokens $\mathbf{S}^{(i)} \in \mathbb{R}^{N_s \times d_{\text{head}}}$ and destination (\texttt{dst}) tokens $\mathbf{D}^{(i)} \in \mathbb{R}^{N_d \times d_{\text{head}}}$. Then, we compute the cosine similarity matrix $\mathbf{Sim}^{(i)} \in \mathbb{R}^{N_s \times N_d}$ between $\mathbf{S}^{(i)}$ and $\mathbf{D}^{(i)}$:

\begin{equation}
\mathbf{Sim}^{(i)} 
= \operatorname{RowNorm}\!\left(\mathbf{S}^{(i)}\right)
\cdot
\operatorname{RowNorm}\!\left(\mathbf{D}^{(i)}\right)^{\top}
\label{eq:sim}
\end{equation}

\noindent For each \texttt{src} token, we only keep the similarity score to its most similar \texttt{dst} token, which is considered its best match. Let $r=N-M$, we merge the top-$r$ \texttt{src} tokens with the highest similarity scores into their best-matching \texttt{dst} token. Then, we compute $r$ \emph{merged tokens} as the mean of all of their \emph{constituent tokens} (the matched \texttt{dst} tokens and all the \texttt{src} tokens that match it).

This merging process is performed separately for the head-wise queries and keys to obtain the merged tokens:
\begin{equation}
    \tilde{\mathbf{Q}}^{(i)} = \mathcal{M}_i^q(\mathbf{Q}^{(i)}), \quad \tilde{\mathbf{K}}^{(i)} = \mathcal{M}_i^k(\mathbf{K}^{(i)}) \,.
\end{equation}
The values $\mathbf{V}^{(i)}$ are merged using the same matches computed in $\mathcal{M}_i^k(\mathbf{K}^{(i)})$, from which we then obtain $\tilde{\mathbf{V}}^{(i)}$. We can now efficiently perform the attention computation on the reduced-length $\tilde{\mathbf{Q}}^{(i)}$/$\tilde{\mathbf{K}}^{(i)}$/$\tilde{\mathbf{V}}^{(i)}$ tokens:
\begin{equation}
    \mathbf{A}^{(i)} = \text{softmax}\left(\frac{\tilde{\mathbf{Q}}^{(i)} (\tilde{\mathbf{K}}^{(i)})^\top}{\sqrt{d_{\text{head}}}}\right) \in \mathbb{R}^{M \times M} \,, 
\end{equation}
\begin{equation}
    \tilde{\mathbf{O}}^{(i)} = \mathbf{A}^{(i)} \tilde{\mathbf{V}}^{(i)} \in \mathbb{R}^{M \times d_{\text{head}}} \,.
\end{equation}

\paragraph{Unmerging}

Finally, we need to restore the full token sequence through a head-wise unmerging step, formalized as a set of functions 
$\mathcal{U}_i: \mathbb{R}^{M \times d_{\text{head}}} \to \mathbb{R}^{N \times d_{\text{head}}}$. Here, we utilize the simple yet effective unmerging procedure proposed in ToMeSD~\cite{bolya2023tomesd}: For each query token $\mathbf{q}_n^{(i)} \in \mathbf{Q^{(i)}}$ from the original query sequence that contributed to a merged query token $\tilde{\mathbf{q}}_m^{(i)} \in \tilde{\mathbf{Q}}^{(i)}$, its final output $\mathbf{o}_n^{(i)}$ is simply the copied output of the merged token $\tilde{\mathbf{o}}_m^{(i)}$:
\begin{equation}
    \mathbf{o}_n^{(i)} := \tilde{\mathbf{o}}_m^{(i)} \,.
\end{equation}
Tokens that were not merged retain their output. 
As a final step, we concatenate the unmerged outputs across all attention heads on the channel dimension to obtain:
\begin{equation}
    \mathbf{O}^{(i)} = \mathcal{U}_i(\tilde{\mathbf{O}}^{(i)}) \in \mathbb{R}^{N \times d_{\text{head}}}\,,
\end{equation}
\begin{equation}
    \mathbf{O} = \operatorname{Concat}(\mathbf{O}^{(1)}, \ldots, \mathbf{O}^{(h)}) \in \mathbb{R}^{N \times d} \,.
\end{equation}
Note that while the queries and keys can make independent merging decisions per head, the unmerging is determined solely by the matches in $\mathcal{M}_i^q(\mathbf{Q}^{(i)})$, because the order of $\tilde{\mathbf{O}}^{(i)}$ is determined by $\tilde{\mathbf{Q}}^{(i)}$. 

\subsection{Temporal Reordering and Merging}
\label{sec:merge}
\begin{figure*}[htbp]
  \centering
  \begin{minipage}{0.7\textwidth}
    \centering
    \begin{subfigure}[t]{0.425\linewidth}
      \includegraphics[width=\linewidth]{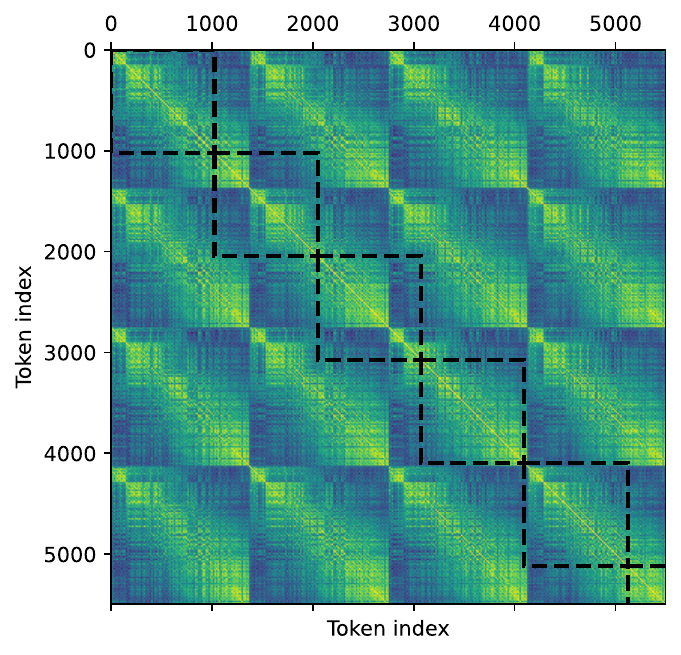}
      \caption{}
      \label{fig:wo_reordering}
    \end{subfigure}
    \hfill
    \begin{subfigure}[t]{0.565\linewidth}
      \includegraphics[width=\linewidth]{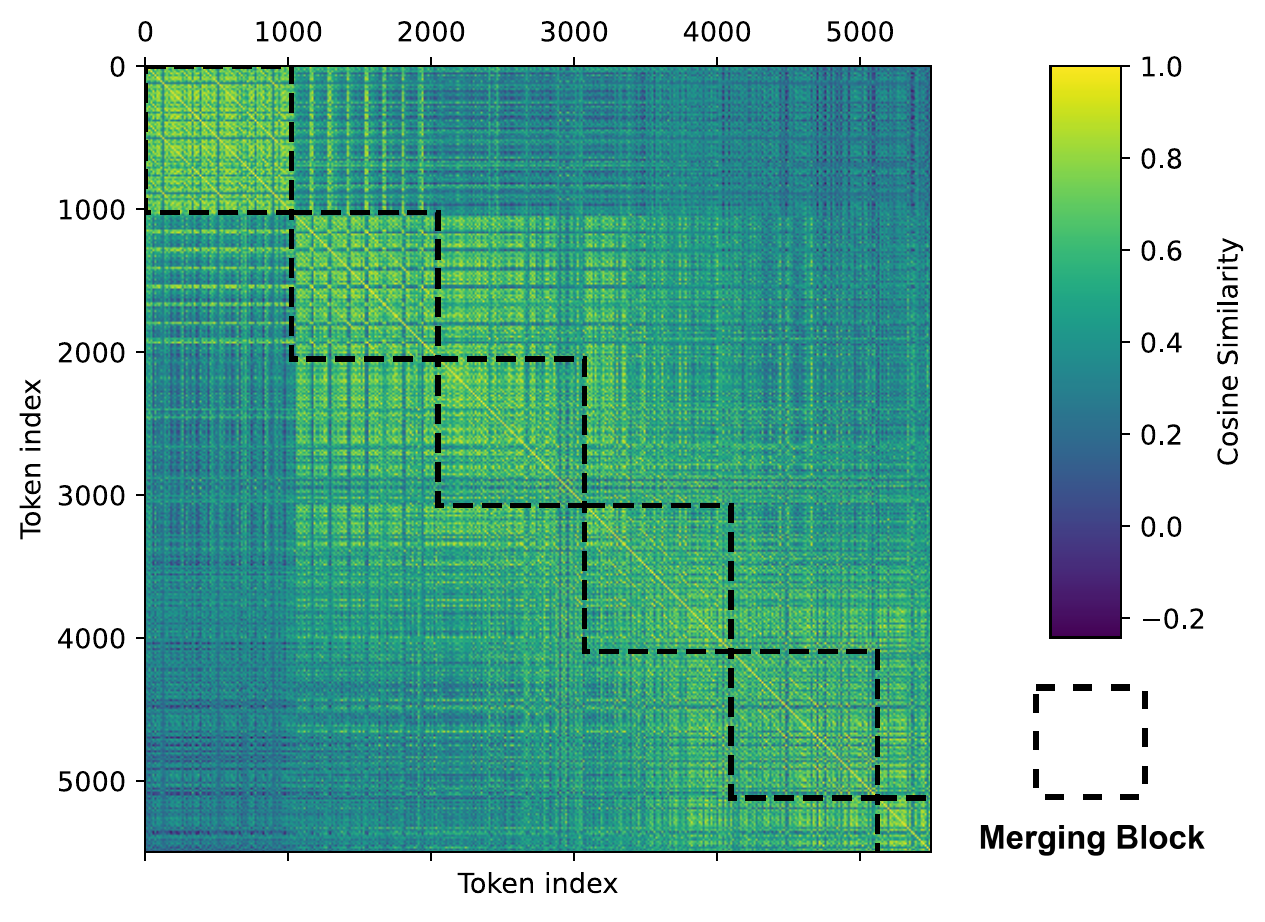}
      \caption{}
      \label{fig:w_reordering}
    \end{subfigure}
    \caption{Token similarity in merging blocks of size 1024. (a) Without temporal reordering, many highly similar matches lie outside of merging blocks. We can't capture those matches unless we use a global merging block that is very costly. (b) Through temporal reordering, high-similarity matches shift inside merging blocks, leading to better merging quality.}
    \label{fig:temporal_merging}
    \vspace{-0.1cm}
  \end{minipage}
\end{figure*}

As stated in Sec.~\ref{sec:tomesd}, the main computational bottleneck of token merging is the computation of the similarity matrix in Eq.~\ref{eq:sim}. To merge $N$ tokens of dimension $d_{\text{head}}$, if we separate them into 25\% \texttt{dst} tokens and 75\% \texttt{src} tokens, the computation of the similarity matrix between them would require around $0.19N^2d_{\text{head}}$ FLOPs, increasing quadratically with the number of tokens. 


To address this challenge, we propose to split the whole token sequence into \emph{merging blocks} of \emph{fixed} size $n_b$ and only perform the token merging within these blocks. In this way, the merging cost grows linearly with $N$. As stated in Sec.~\ref{sec:tomesd}, we merge the top-$r$ \texttt{src} based on the exact similarity matrix. This implies three key points\footnote{A proof of these statements can be found in the Appendix.} to formulate a good block-wise token merging strategy:

\begin{itemize}
    \item The block similarity matrix computed between tokens in the merging block is composed of similarity scores from the global all-to-all similarity matrix.
    \item The quality of the block-wise token merging method is dependent on the number of high similarity scores that we can include in the block similarity matrix.
    \item Given a splitting strategy, larger merging block sizes are always better at the expense of higher merging costs. 
\end{itemize}


Hence, the challenge is how to split the token sequence into merging blocks of a given size to maximize the number of high similarity scores in the block similarity matrix.

Given our findings in Sec.~\ref{sec:observationsandinsights}, highly similar tokens reside in neighborhoods, implying that - given the query tokens of a head $\mathbf{Q}^{(i)}\in \mathbb{R}^{N \times d_{\text{head}}}$ - if we merge these tokens in consecutive merging blocks along the $N$ dimension, we can capture the spatial similarity along the main diagonal of the similarity matrix. The negative implication is that we will lose track of many highly similar matches outside the merging block. In Fig.~\ref{fig:wo_reordering}, we visualize the local similarity matrix of merging blocks of size 1024 on the global similarity matrix of 4 frames with 1374 tokens each. It can be observed that a lot of high similarity scores fall outside of the merging blocks' similarity matrices, making them unconsidered in the merging process. Intuitively, this is because the spatially expanding merging block fails to capture the temporal correspondence we observed in Sec~\ref{sec:observationsandinsights}.

To solve this problem, we propose a simple but effective technique: \emph{Temporal Reordering}. As shown in Fig.~\ref{fig:overview}, before the token merging process, we reorder the tokens so that similar spatial blocks of size $n_s$ (marked with the same color) across $n_t$ frames are stacked together to form temporal merging blocks of size $n_b = n_s\times n_t$ that consist of highly similar tokens. Then, we can perform efficient block-wise token merging in these temporal merging blocks of size $n_b << N$ and reduce the token length of Q/K/V, significantly accelerating the attention computation. After the attention computation, we first unmerge the tokens in temporal blocks to recover their original sizes, then we reorder the tokens to make sure that the order of output tokens aligns with the input tokens.

In Fig.~\ref{fig:w_reordering}, we perform a temporal reordering where we stack spatial blocks of size $n_s=128$ from $n_t=8$ frames. Compared to Fig.~\ref{fig:wo_reordering}, more high similarity scores are included in the merging blocks after the temporal reordering, meaning that more highly-similar token matches can be considered by the merging process, leading to better overall merging quality. The criteria to choose the appropriate size of spatial blocks and number of temporal frames for the reordering and merging are discussed in Sec.~\ref{sec:pareto}.


\begin{table*}[htb]
\centering
\begin{tabular}{lcc|ccc|ccc|cc}
\hline
          &      &      & \multicolumn{3}{c|}{7 Scenes (Stride 10)} 
                         & \multicolumn{3}{c|}{NRGBD (Stride 10)} 
                         & \multicolumn{2}{c}{ETH3D} \\
\hline
          & Q Ratio & K/V Ratio 
          & Acc.$\downarrow$ & Comp.$\downarrow$ & Time$\downarrow$ 
          & Acc.$\downarrow$ & Comp.$\downarrow$ & Time$\downarrow$ 
          & Acc.$\downarrow$ & Comp.$\downarrow$ \\
\hline
CUT3R~\cite{wang2025continuous} 
    & 1.00 & 1.00 
    & 0.041 & 0.029 & 4.2s 
    & 0.132 & 0.056 & 5.7s 
    & / & / \\

VGGT*~\cite{wang2025vggt}      
    & 1.00 & 1.00    
    & 0.019 & 0.021 & 9.1s 
    & 0.010 & 0.010 & 13.9s 
    & 0.087 & 0.053 \\

FastVGGT~\cite{shen2025fastvggt}   
    & 0.34 & 0.34 
    & 0.018 & 0.020 & 4.5s 
    & 0.016 & 0.013 & 7.0s 
    & 0.120 & 0.076 \\ 

\rowcolor{green!15}
VGGT*+HTTM 
    & \textbf{0.20}  & 0.30  
    & 0.020 & 0.023 & 4.3s  
    & 0.012 & 0.010 & 6.8s 
    & 0.091 & 0.054 \\
\hline
\end{tabular}
\caption{Comparison of reconstruction performance in terms of accuracy (Acc.) and completeness (Comp.) measured by Chamfer distance across multiple datasets.}
\label{tab:reconstruction}
\end{table*}

\subsection{Adaptive Outlier Filtering}
\label{sec:outliers}
For better parallelization, we use a fixed spatial block size and number of temporal frames when forming the merging blocks across heads, which does not always align with the similarity pattern across frames and heads. Furthermore, we use the same merging ratio for all these merging blocks. Hence, for merging blocks with low intra-block similarity, low-similarity tokens will be merged, resulting in large distances between the original tokens and the merged token. We call these tokens \emph{outliers}. We filter outliers adaptively to improve the overall representational ability of the merged tokens:

\begin{enumerate}
    \item We perform an initial merging step of query tokens inside the block as described in Sec.~\ref{sec:tomesd}.
    \item Given the merged queries $\tilde{\mathbf{Q}}\in \mathbb{R}^{h\times M\times d_{head}}$ for all heads, we compute the deviation from all the original query tokens to their corresponding merged tokens in $L_2$ distances. For the tokens that are not merged with other tokens, the deviation is set to zero.
    \item We then identify the top $d$\% of tokens with the largest deviations to their merged tokens across \emph{all} heads and mark them as outliers, yielding a binary outlier mask $\mathbf{M}_o \in \{0,1\}^{h \times N}$.
    \item We update the \emph{merged tokens} if any of their \emph{constituent tokens} (as defined in Sec.~\ref{sec:merge}) are marked as outliers. The contributions of the outliers are subtracted from the merged tokens, and the outliers are no longer merged.
\end{enumerate}

\noindent Step 3 provides greater flexibility to heads with more outliers under the same overall budget. Additionally, step 4 protects the representational ability of merged tokens and keeps the uniqueness of outliers. Note that we only apply the outlier filtering on query tokens.

We implement a custom CUDA kernel to realize this filtering logic efficiently through block-wise parallelization and on-the-fly deviation computation. More details about this CUDA kernel are in the Appendix, and the latency overhead is reported in Sec.~\ref{sec:latency}.



\vspace{-0.15cm}
\section{Experiments}
\begin{figure*}[htb]
    \centering
    \begin{subfigure}{0.32\textwidth}
        \includegraphics[width=\linewidth]{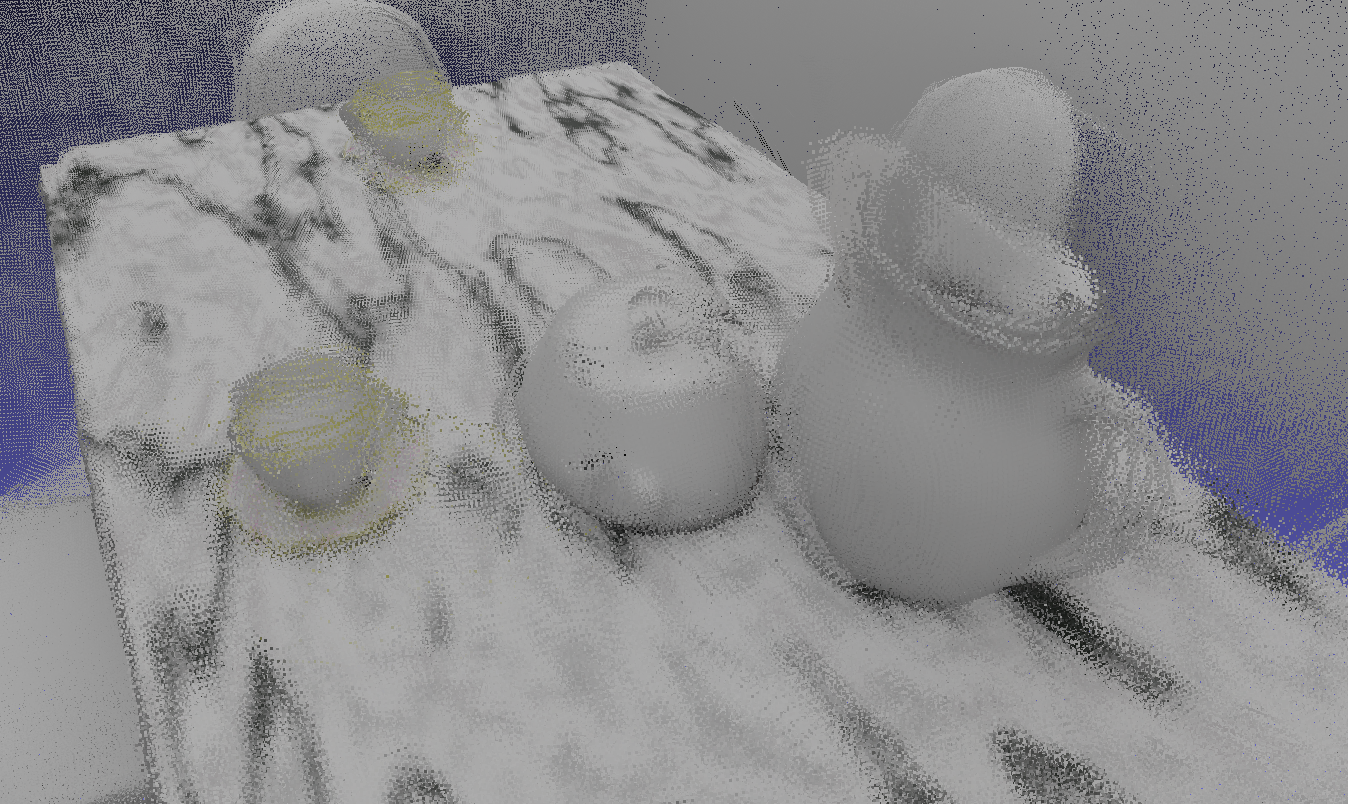}
        \caption{Standard VGGT}
    \end{subfigure}
    \begin{subfigure}{0.32\textwidth}
        \includegraphics[width=\linewidth]{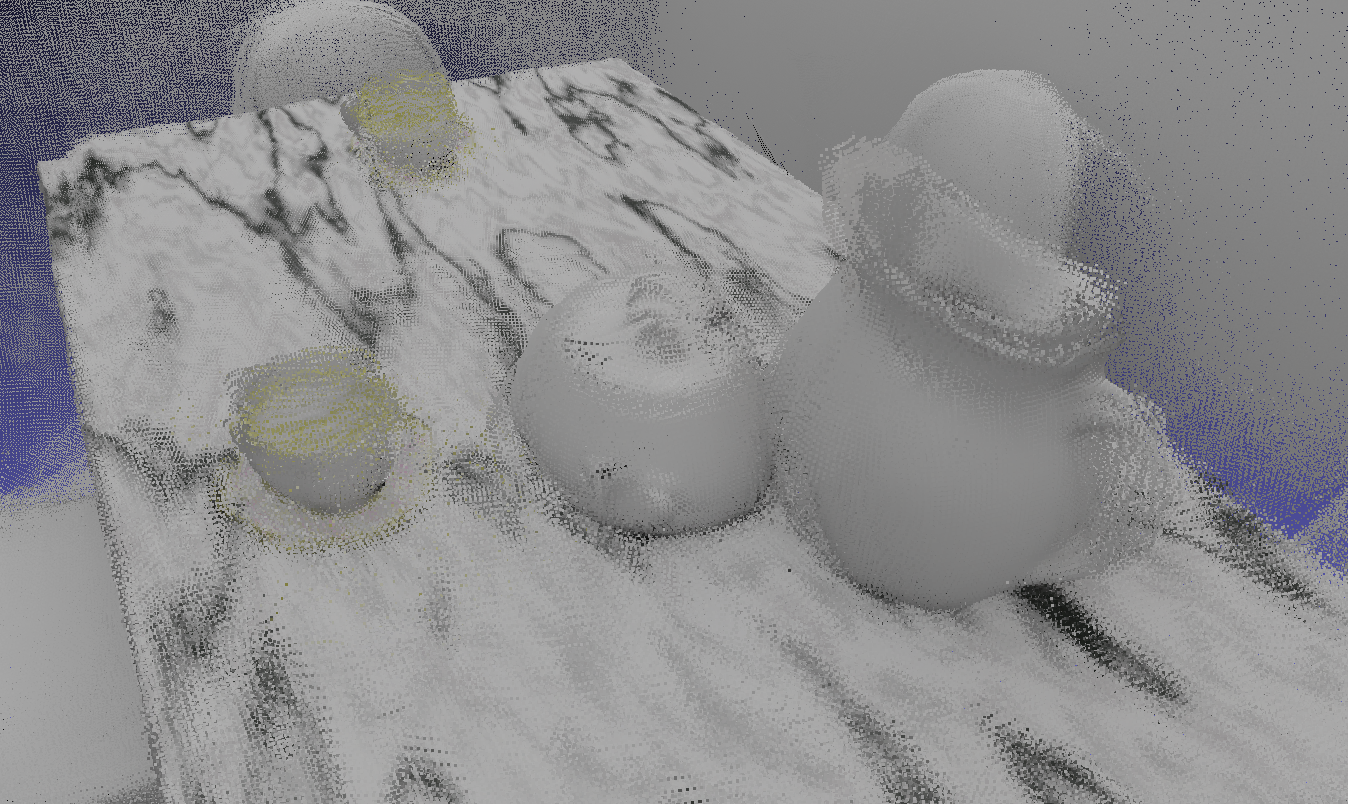}
        \caption{VGGT+HTTM}
    \end{subfigure}
    \begin{subfigure}{0.32\textwidth}
        \includegraphics[width=\linewidth]{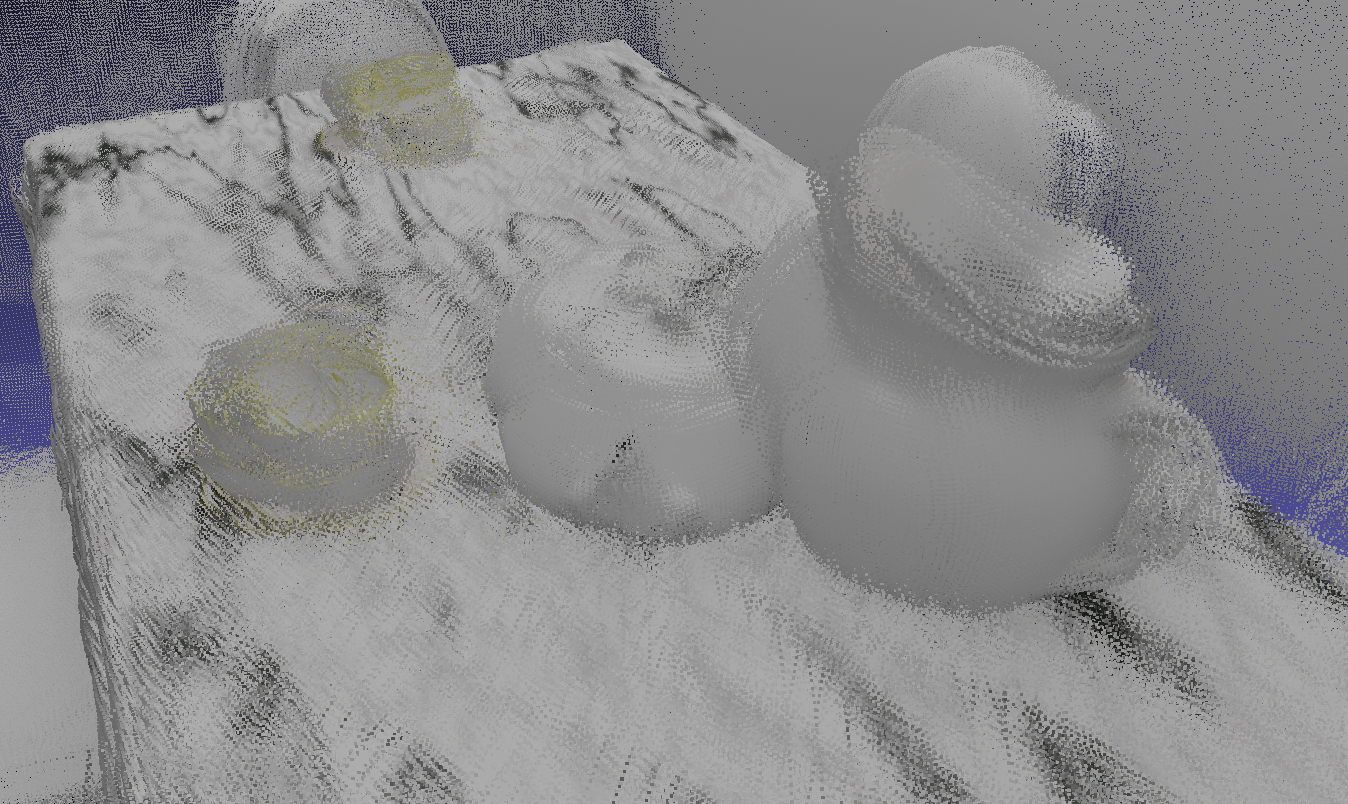}
        \caption{FastVGGT}
    \end{subfigure}
    \caption{Qualitative results on a scene from NRGBD. Compared to FastVGGT, HTTM preserves more high-fidelity details of VGGT.}
\label{fig:quality}
\end{figure*}

\subsection{3D Reconstruction Results}
In this section, we report the quantitative performance of applying HTTM on 7Scenes~\cite{shotton2013scene} and NRGBD~\cite{azinovic2022neural} to reduce the token sequence length of VGGT.
\label{sec:3d_recon}

\noindent \textbf{Experiment Setup}\quad For the temporal reordering described in Sec.~\ref{sec:merge}, we use a spatial block size of $n_s=128$ and a temporal frame length of $n_t=30$ to form temporal merging blocks of size $n_b=3840$. We merge 90\% of the query tokens and filter 10\% outliers as stated in Sec.~\ref{sec:outliers}, so that in total, the reduced query sequence is 20\% of its original length. For key and value tokens, we use a 70\% merging ratio, resulting in 30\% of the original length. For FastVGGT, we use its standard merging setup, which results in around 34\% actual sequence length. We didn't further increase its merging ratios because FastVGGT already underperforms HTTM on NRGBD in this setup.

The reported reconstruction results use the depth and camera head of VGGT since they yield better results, as stated in the original VGGT paper\cite{wang2025vggt}. For the baseline, we use the VRAM-efficient VGGT implemented in FastVGGT\cite{shen2025fastvggt}, denoted as VGGT*, which discards unused intermediate outputs during inference \textbf{without} affecting reconstruction quality. All inferences are performed in Bfloat16\cite{wang2019bfloat16} using FlashAttention\cite{dao2023flashattention} on an Nvidia A100.

\noindent \textbf{Results}\quad
As shown in Table~\ref{tab:reconstruction}, we evaluate the 3D reconstruction performance on 7Scenes and NRGBD with keyframes sampled every 10 frames. Compared to the baseline VGGT, HTTM maintains comparable performance with much shorter Q/K/V sequence length. Compared to FastVGGT, HTTM surpasses FastVGGT on NRGBD and achieves comparable results on 7Scenes using smaller sequence lengths (higher merging ratios). As shown in Fig.~\ref{fig:quality}, HTTM preserves more high-fidelity details of the reconstruction result from the original VGGT. We further evaluate on ETH3D~\cite{schoeps2017cvpr} to show HTTM's performance on weakly overlapped inputs. More evaluations on longer input sequences can be found in the Appendix.


\begin{table*}[htb]
\centering
\begin{minipage}{0.58\textwidth}
\centering
\begin{tabular}{l|cc|c|c|c|c}
\hline
            & \multicolumn{2}{c|}{Token Ratio} & \multicolumn{4}{c}{Number of Frames} \\ \hline
            & Q      & K/V    & 100  & 300  & 500  & 1000 \\
\hline
VGGT*      & 1.00 & 1.00 & 9.1s & 60.7s & 177.5s & 724.6s \\
\hline
Fast VGGT  & 0.34 & 0.34 & 4.5s & 22.4s & 52.3s & 175.2s \\
\hline
VGGT*+HTTM & 0.4  & 0.3  & 4.4s & 18.6s & 38.0s & 130.0s \\
\hline
\rowcolor{green!15}
VGGT*+HTTM & 0.2  & 0.3  & 4.3s & 16.3s & 35.8s & \textbf{102.8s} \\
\hline
\end{tabular}
\caption{Latency Comparison. At 1000 frames, we are \textbf{7}$\times$ faster than the baseline VGGT with FlashAttention in Bfloat16.}
\label{table:latency}
\end{minipage}
\hfill
\begin{minipage}{0.38\textwidth}
\centering
\begin{tabular}{l|cc}
\hline
                    & HTTM  & FastVGGT \\ \hline
Attention Kernel    & 2.95s & 2.97s     \\ \hline
\rowcolor{green!15}
Matching            & 0.12s & 2.31s     \\ \hline
Merged Token Aggr. & 0.41s & 0.11s    \\ \hline
\end{tabular}
\caption{Averaged latency composition of Global Attention layers in HTTM and FastVGGT using comparable merging ratios over 1000 frames.}
\label{table:composition}
\end{minipage}
\vspace{-0.4cm}
\end{table*}

\subsection{Latency}
\label{sec:latency}
In this section, we evaluate the acceleration ability of HTTM for long input sequences. The experiment setup is the same as in Sec.~\ref{sec:3d_recon}. We compare the latency of HTTM to FastVGGT under similar merging ratios and comparable (or better) task performances in Table~\ref{tab:reconstruction}. 

\noindent \textbf{Under similar token merging ratios}\quad HTTM shows better latency performance due to our block-wise token merging design after temporal reordering as described in Sec.~\ref{sec:merge}. As shown in Table~\ref{table:composition}, the latency for executing the attention computation is similar for HTTM and FastVGGT when using comparable merging ratios. However, by only performing token merging within temporal merging blocks, HTTM largely reduces the matching cost, leading to the overall latency improvement over FastVGGT in Table~\ref{table:latency}. Due to our adaptive outlier filtering design introduced in Sec.~\ref{sec:outliers}, the merged token aggregation latency of HTTM is higher, but the overhead is acceptable compared to the matching cost reduction. Combine the matching and aggregation overhead together, HTTM achieves 4.58$\times$ latency reduction on the merging cost as shown in Fig.~\ref{fig:intro_fig}.

\noindent \textbf{Under comparable task performance}\quad 
HTTM is able to achieve comparable task performance to the baseline VGGT and FastVGGT using more drastic merging ratios as shown in Sec.~\ref{sec:3d_recon}. Hence, we report the latency performance over long sequences under this configuration. As shown in Table~\ref{table:latency}, in this setup, HTTM further reduces the latency. With 1000 input frames, HTTM substantially reduces the latency by 7$\times$ compared to the baseline VGGT.

\subsection{Spatial Merging vs.\ Temporal Merging}
\label{sec:pareto}
In this section, we want to explore the effectiveness of spatial merging versus temporal merging. To do that, we first define the \emph{merging cost} (as discussed in Sec.~\ref{sec:merge}) as the merging block size $n_b = n_s\times n_t$, where $n_s$ is the spatial block size and $n_t$ is the temporal frame length during the temporal reordering. The \emph{merging quality} is defined as the 10th quantile of the similarities between \emph{merged} matches.

\begin{figure}[thbp]
  \centering
    \centering
    \begin{subfigure}[t]{0.478\linewidth}  
      \includegraphics[width=\linewidth]{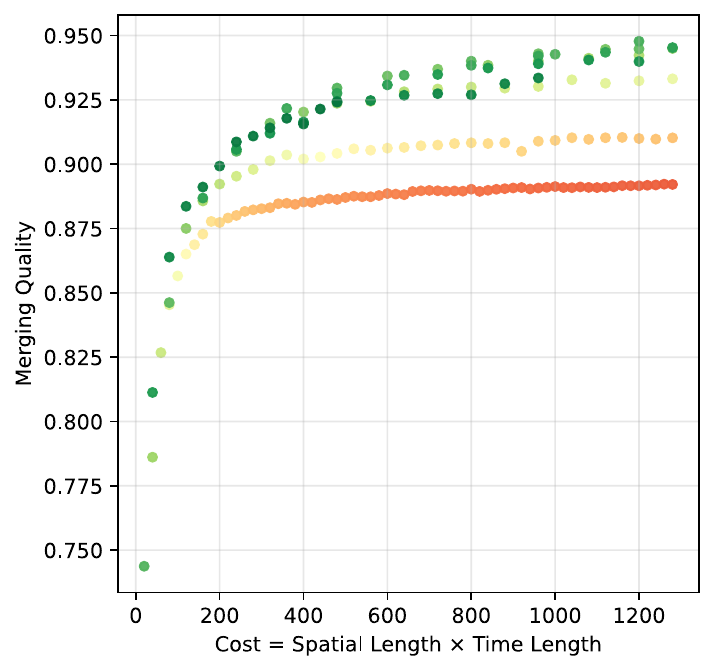}
      \caption{Continuous Frames}
      \label{fig:pareto_a}
    \end{subfigure}
    \hfill
    \begin{subfigure}[t]{0.502\linewidth}  
      \includegraphics[width=\linewidth]{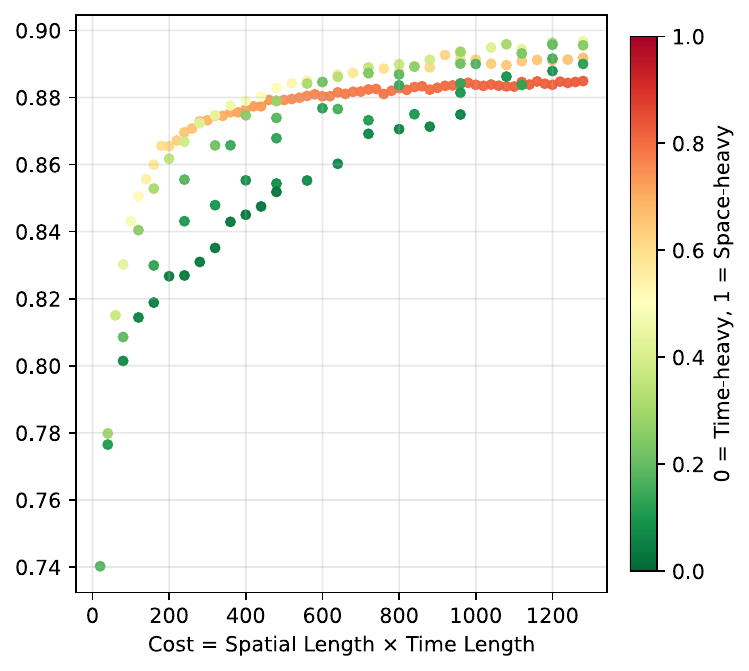}
      \caption{Sparse-view Frames}
      \label{fig:pareto_b}
    \end{subfigure}
    \caption{The Pareto front illustrates the trade-off between merging cost and merging quality, with color indicating the composition of the cost. \textcolor[rgb]{0,0.3,0}{Greenish} points merge more frames along the temporal dimension, while \textcolor[rgb]{0.5,0,0}{redish} points merge more tokens along the spatial dimension.}
    \label{fig:pareto}
\vspace{-0.3cm}
\end{figure}

In Fig.~\ref{fig:pareto}, we visualize the Pareto front between the merging cost and merging quality on two scenes, one with highly similar, temporally continuous frames, and one with sparse-view frames. Additionally, we visualize the cost composition with colors. Given a fixed cost $n_b$, points with larger $n_s$ will become reddish, and points with larger $n_t$ will become greenish.

It can be observed that in both cases, merging quality grows with the merging cost. For continuous frames (Fig.~\ref{fig:pareto_a}), merging predominantly along the temporal dimension yields better performance, which aligns with our observation in Sec.~\ref{sec:observationsandinsights}. For sparse-view frames (Fig.~\ref{fig:pareto_b}), merging mainly along the spatial dimension becomes more effective. However, when given a higher cost budget ($\geq$ 800), it is still beneficial to perform temporal merging (green and yellow points) rather than solely performing spatial merging (red points).
\section{Ablation Study}
\subsection{Adaptive Outlier Filtering}
In this section, we show the necessity of our adaptive outlier filtering. We evaluate HTTM on the NRGBD under three merging setups for the query tokens. For the key and value tokens, we merge 70\% of them in both cases.

\begin{itemize}
    \item Merging 90\% query tokens with 10\% outlier filtering.
    \item Merging 85\% query tokens with 5\% outlier filtering.
    \item Merging 80\% query tokens.
\end{itemize}

\noindent These three setups result in the same number of Q/K/V tokens after merging. As shown in Table~\ref{tab:ablation}, although the three setups use the same token length, not performing outlier filtering leads to a catastrophic performance drop, showing the necessity of outlier filtering in block-wise token merging as discussed in Sec.~\ref{sec:outliers}.

\begin{table}[h]
\centering
\begin{tabular}{l|ll}
\hline
                            & Acc.$\downarrow$ & Comp.$\downarrow$ \\ \hline
Without outlier filtering   & 0.240   &  0.310   \\ \hline
With 5\% outlier filtering  & 0.013   &  0.011   \\ \hline
With 10\% outlier filtering & 0.012   &  0.010   \\ \hline
\end{tabular}
\caption{Accuracy and completeness on NRGBD with and without outlier filtering using the same token sequence length.}
\label{tab:ablation}
\end{table}

\vspace{-0.6cm}
\section{Conclusion}
In this work, we propose HTTM, a training-free token merging approach that accelerates VGGT's inference. We conduct systematic explorations of similarity patterns in VGGT and analyze the main limitations of existing methods in merging efficiency and representational ability. To leverage the spatial locality and temporal correspondence of VGGT's tokens at the head-embedding level, we introduce a temporal reordering and head-wise adaptive outlier filtering technique that helps HTTM merge tokens efficiently while preserving their uniqueness, leading to substantial acceleration up to \textbf{7}$\times$ over the original VGGT on long input sequences without performance degradation.
{
    \small
    \bibliographystyle{ieeenat_fullname}
    \bibliography{main}
}

\clearpage
\setcounter{page}{1}
\maketitlesupplementary

\appendix

\section{RoPE's Effect on the Similarity Pattern}

\begin{figure*}[htbp]
\centering
\begin{tabular}{cccc}
\begin{subfigure}{0.23\textwidth}
    \centering
    \includegraphics[width=\linewidth]{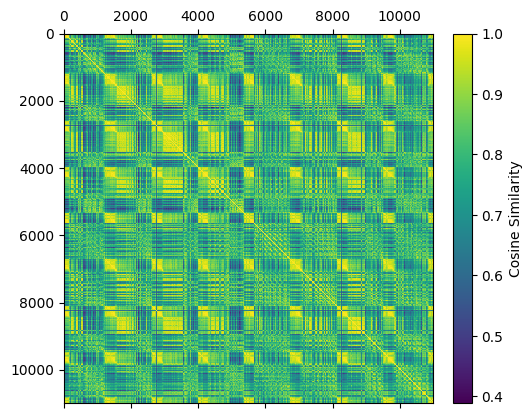}
    \caption{Layer 0 before RoPE}
    \label{fig:layer0_bf}
\end{subfigure} &
\begin{subfigure}{0.23\textwidth}
    \centering
    \includegraphics[width=\linewidth]{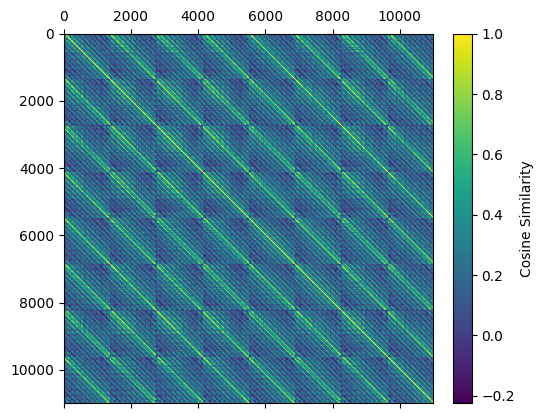}
    \caption{Layer 0 after RoPE}
    \label{fig:layer0_aft}
\end{subfigure} &
\begin{subfigure}{0.23\textwidth}
    \centering
    \includegraphics[width=\linewidth]{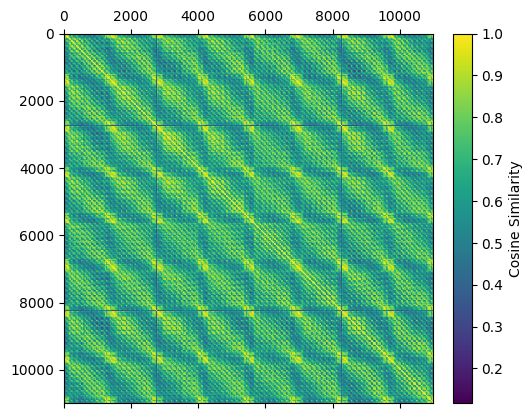}
    \caption{Layer 1 before RoPE}
\end{subfigure} &
\begin{subfigure}{0.23\textwidth}
    \centering
    \includegraphics[width=\linewidth]{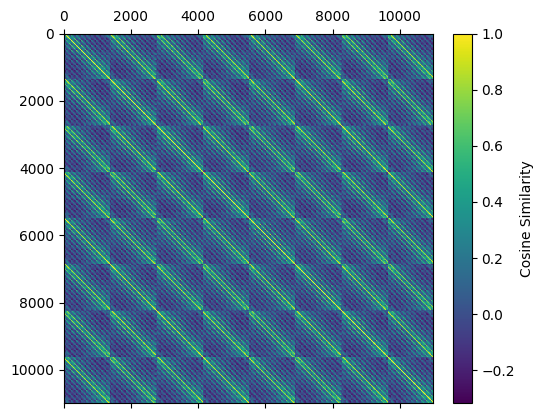}
    \caption{Layer 1 after RoPE}
\end{subfigure} \\

\begin{subfigure}{0.23\textwidth}
    \centering
    \includegraphics[width=\linewidth]{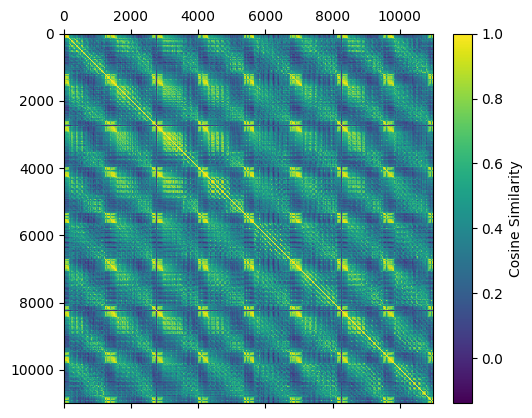}
    \caption{Layer 2 before RoPE}
\end{subfigure} &
\begin{subfigure}{0.23\textwidth}
    \centering
    \includegraphics[width=\linewidth]{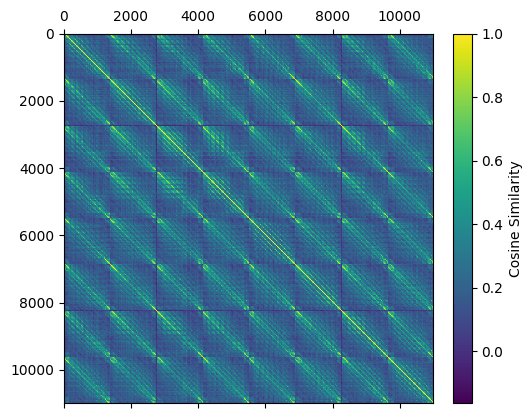}
    \caption{Layer 2 after RoPE}
\end{subfigure} &
\begin{subfigure}{0.23\textwidth}
    \centering
    \includegraphics[width=\linewidth]{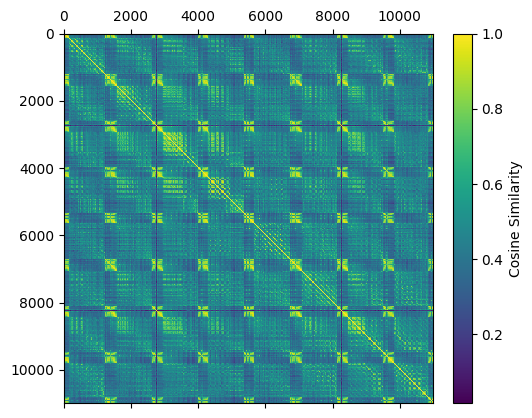}
    \caption{Layer 3 before RoPE}
\end{subfigure} &
\begin{subfigure}{0.23\textwidth}
    \centering
    \includegraphics[width=\linewidth]{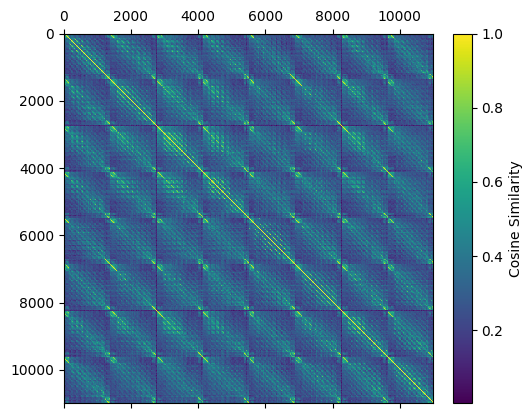}
    \caption{Layer 3 after RoPE}
\end{subfigure}
\end{tabular}
\caption{Query token similarity maps before and after RoPE in Frame Attention layers with non-overlapping input frames}
\label{fig:rope}
\end{figure*}

\begin{figure*}[htbp]
\centering
\begin{tabular}{cccc}
\begin{subfigure}{0.23\textwidth}
    \centering
    \includegraphics[width=\linewidth]{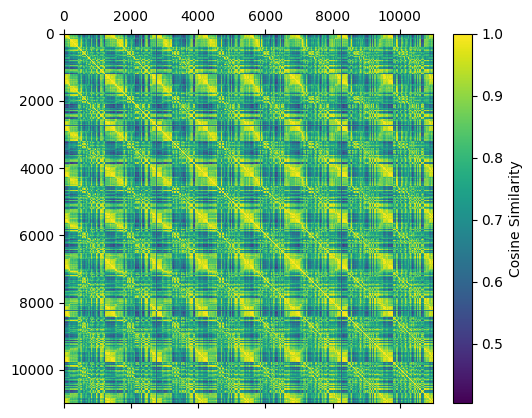}
    \caption{Layer 0 before RoPE}
\end{subfigure} &
\begin{subfigure}{0.23\textwidth}
    \centering
    \includegraphics[width=\linewidth]{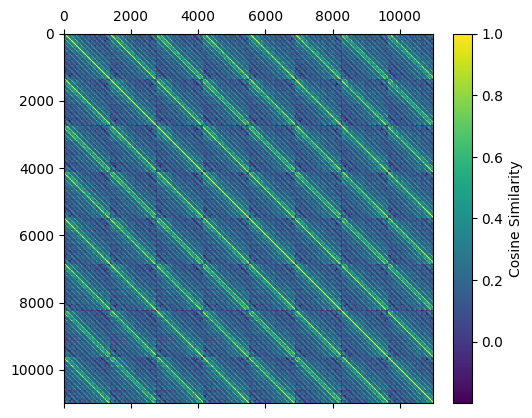}
    \caption{Layer 0 after RoPE}
\end{subfigure} &
\begin{subfigure}{0.23\textwidth}
    \centering
    \includegraphics[width=\linewidth]{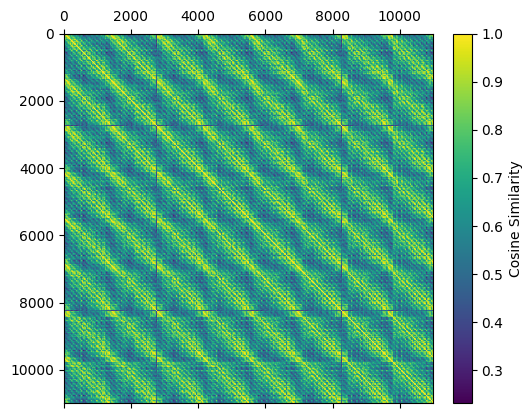}
    \caption{Layer 1 before RoPE}
\end{subfigure} &
\begin{subfigure}{0.23\textwidth}
    \centering
    \includegraphics[width=\linewidth]{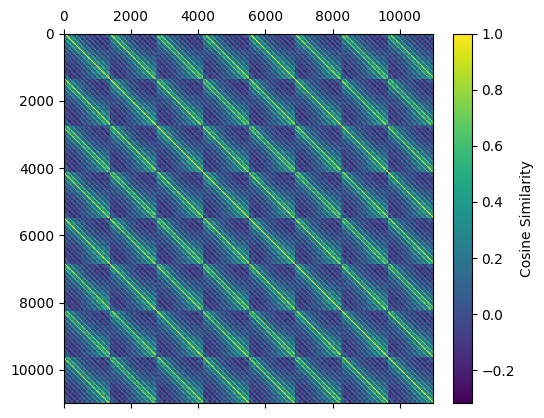}
    \caption{Layer 1 after RoPE}
\end{subfigure} \\

\begin{subfigure}{0.23\textwidth}
    \centering
    \includegraphics[width=\linewidth]{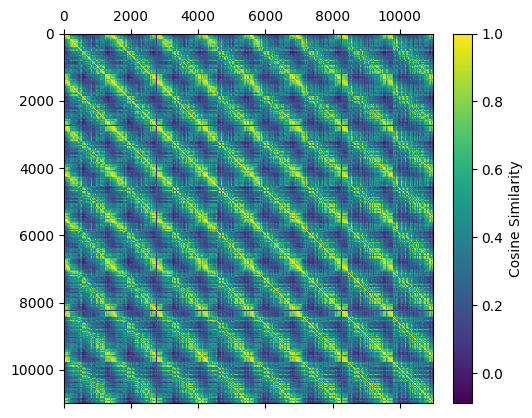}
    \caption{Layer 2 before RoPE}
\end{subfigure} &
\begin{subfigure}{0.23\textwidth}
    \centering
    \includegraphics[width=\linewidth]{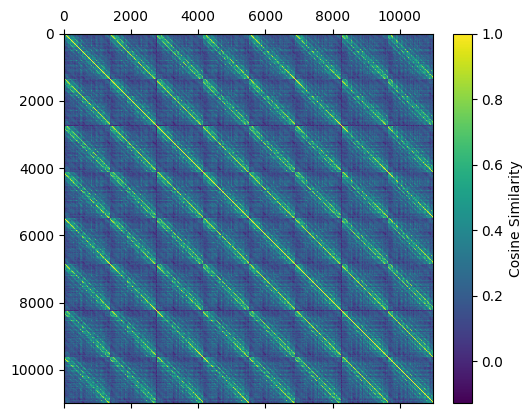}
    \caption{Layer 2 after RoPE}
\end{subfigure} &
\begin{subfigure}{0.23\textwidth}
    \centering
    \includegraphics[width=\linewidth]{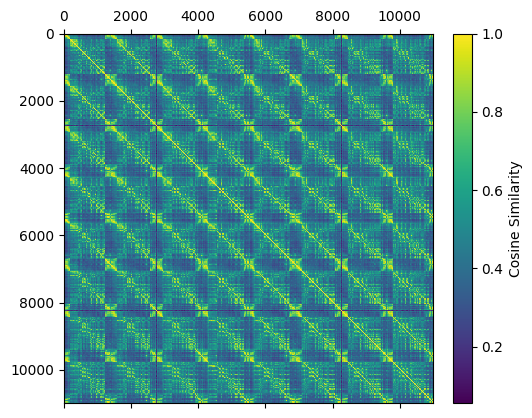}
    \caption{Layer 3 before RoPE}
\end{subfigure} &
\begin{subfigure}{0.23\textwidth}
    \centering
    \includegraphics[width=\linewidth]{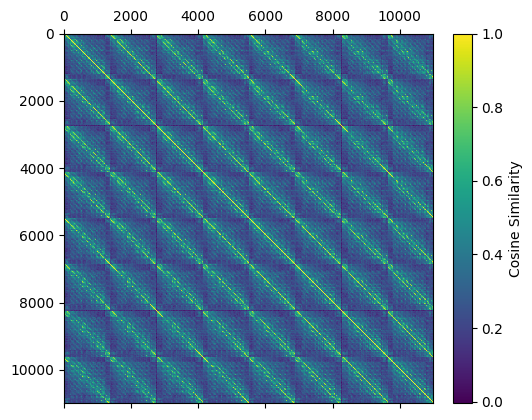}
    \caption{Layer 3 after RoPE}
\end{subfigure}
\end{tabular}
\caption{Query token similarity maps before and after RoPE in Frame Attention layers with temporally continuous input frames.}
\label{fig:rope_continuous}
\end{figure*}

In this section, we investigate the Rotary Position Embedding (RoPE~\cite{su2024roformer})'s effect on the strong periodic patterns observed in Fig.~\ref{fig:token_sim}. To show that the high similarity values near the off-diagonals do emerge from the RoPE, in Fig.~\ref{fig:rope}, we visualize the similarity map of query tokens in the early Frame Attention layers with \textbf{non-overlapping} input frames. It can be observed that the input feature of the first Frame Attention layer (DINO features) does not exhibit high temporal values near off-diagonals (Fig.~\ref{fig:layer0_bf}), which align with the non-overlapping input frames. However, after applying the frame-wise RoPE for the first time, high similarity values near the off diagonals emerge as shown in Fig.~\ref{fig:layer0_aft}. After that, it can be observed that, in each layer, the spatial distinctiveness within frames is enhanced after applying the frame-wise RoPE.

In Fig.~\ref{fig:rope_continuous}, we visualize the similarity maps with temporally continuous input frames. With these inputs, the changes in similarity patterns before and after applying RoPE are similar, but high similarity values near the off-diagonals are more vivid.

\section{Theoretical Proofs For Block-Wise Token Merging}

We provide proofs for the three statements made in Sec.~\ref{sec:merge}. For clarity, we omit the head index $i$ in this section.
Let the global source and destination token sets be $\mathcal{S}$ and $\mathcal{D}$, $\mathcal{S}\cap\mathcal{D}=\emptyset$.
The entries of the global similarity matrix $W\in\mathbb{R}^{|\mathcal{S}|\times|\mathcal{D}|}$ is defined as:
\[
W_{ij} = \mathrm{sim}(s_i,d_i), \quad s_i \in \mathcal{S},d_j \in \mathcal{D}
\]

\noindent For a merging budget $r$, the merging rule selects the \emph{top-$r$} source--destination pairs with the largest similarities.  
For any selected set $\mathcal{M}$ of merging candidates, the merging quality $Q$ is defined as the \emph{average} similarity between merged matches:
\[
Q(\mathcal{M}):=\frac{1}{r}\sum_{(s,d)\in\mathcal{M}} \mathrm{sim}(s,d).
\]

\noindent In block-wise token merging, we partition the tokens into $K$ disjoint blocks $\{\mathcal{B}_1,\dots,\mathcal{B}_K\}$. Assuming the same splitting strategy, the source and destination token sets $\mathcal{S}_k$ and $\mathcal{D}_k$ inside block $k \in \{1,\dots,K\}$ are:
\[
\mathcal{S}_k=\mathcal{S}\cap\mathcal{B}_k,\qquad
\mathcal{D}_k=\mathcal{D}\cap\mathcal{B}_k,
\]
After establishing the notations, we proceed to prove the aforementioned statements.
\subsection*{1. Block similarity matrices are submatrices of the global matrix}

\paragraph{Prop.}
For every block $\mathcal{B}_k$, its block-wise similarity matrix $W^{(k)}$ is a submatrix of $W$.

\paragraph{Proof.}
For each $\mathcal{B}_k$, the entries of its similarity matrix is defined as:
\[
W^{(k)}_{i,j}=\mathrm{sim}(s_i,d_j),\qquad s_i\in\mathcal{S}_k,\ d_j\in\mathcal{D}_k.
\]
Since $\mathcal{S}_k \subseteq \mathcal{S}, \mathcal{D}_k \subseteq \mathcal{D}$, it follows that $s_i\in\mathcal{S},\ d_j\in\mathcal{D}$. Therefore, each entry $\mathrm{sim}(s_i,d_j)$ is also a entry in $W$.

\subsection*{2. Merging quality depends on how many high-similarity pairs fall inside blocks}

\paragraph{Prop.}
Let $\mathcal{E}=\mathcal{S}\times\mathcal{D}$ be all possible source--destination pairs, and let
\[
\mathcal{E}_{\mathrm{blk}}:=\bigcup_{k=1}^K \left(\mathcal{S}_k\times\mathcal{D}_k\right)
\]
be the set of pairs permitted by block-wise merging.  
If more large entries of $W$ lie inside $\mathcal{E}_{\mathrm{blk}}$, then the block-wise merging quality increases.

\paragraph{Proof.}
As stated in Sec.~\ref{sec:tomesd}, we pick the top-$r$ best matches with the highest similarity, which yields the global optimal merging quality:

\[\begin{aligned}
\mathcal{M}^\star
&= \arg\max_{\mathcal{M}} \mathrm{sim}(s,d) = \arg\max_{\mathcal{M}} Q(\mathcal{M}) \\
\text{s.t.}\;\; & (s,d)\subset \mathcal{M},\quad
\mathcal{M}\subseteq \mathcal{E},\quad
|\mathcal{M}| = r .
\end{aligned}\]

\noindent Block-wise merging is constrained to subsets of $\mathcal{E}_{\mathrm{blk}}$:
\[\begin{aligned}
\mathcal{M}_{blk}^\star
&= \arg\max_{\mathcal{M}} \mathrm{sim}(s,d) \\
\text{s.t.}\;\; & (s,d)\subset \mathcal{M},\quad
\mathcal{M}\subseteq \mathcal{E}_{blk},\quad
|\mathcal{M}| = r .
\end{aligned}\]
Since $\mathcal{E}_{\mathrm{blk}}\subseteq\mathcal{E}$, the feasible set of block-wise solutions is smaller, hence
\[
Q(\mathcal{M}_{\mathrm{blk}}^\star)\le Q(\mathcal{M}^\star).
\]
Define $H_{blk}$ as the number of optimal merging candidates included in $\mathcal{M}_{blk}^\star$:
\[
H_{blk} := |\mathcal{M}_{blk}^\star \cap \mathcal{M}^\star|
\]
If $H_{blk} < r$, then $Q(\mathcal{M}_{blk}^\star)$ is strictly smaller than $Q(\mathcal{M^\star})$. By including more $s,d$ pairs with high similarity in $\mathcal{E}_{blk}$, we can only increase $H_{blk}$. Therefore, block-wise merging quality is monotone in the number of high-similarity entries of $W$ located inside the blocks.

\subsection*{3. Larger blocks improve merging quality but require more computations}

\paragraph{Prop.}
Fix a splitting strategy that forms blocks of size $n_b$.  
If the block size increases, then (i) the achievable merging quality does not decrease, and (ii) the computational cost grows approximately linearly in $n_b$.

\paragraph{Proof.}

\textit{(i) Larger blocks improve or maintain quality.}

Let two block sizes $n_b^{(1)} < n_b^{(2)}$ be given.  
Under the same splitting strategy, every small block is contained in a unique larger block. Hence
\[
\mathcal{E}_{\mathrm{blk}}^{(1)}
\subseteq
\mathcal{E}_{\mathrm{blk}}^{(2)}.
\]
Thus, the feasible merging sets under smaller blocks are a subset of those under the larger blocks, so
\[
\max_{\substack{\mathcal{M}\subseteq\mathcal{E}_{\mathrm{blk}}^{(1)}\\|\mathcal{M}|=r}} Q(\mathcal{M})
\ \le\
\max_{\substack{\mathcal{M}\subseteq\mathcal{E}_{\mathrm{blk}}^{(2)}\\|\mathcal{M}|=r}} Q(\mathcal{M}).
\]
Therefore, the optimal block-wise average similarity is non-decreasing in $n_b$.

\vspace{1em}
\begin{table*}[ht]
\centering
\begin{tabular}{lcc|ccc|ccc|ccc}
\hline
          &      &      & \multicolumn{3}{c|}{NRGBD (Stride 3)} 
                         & \multicolumn{3}{c|}{ScanNet (500 Frames)}
                         & \multicolumn{3}{c}{ScanNet (1000 Frames)} \\
\hline
          & Q Ratio & K/V Ratio 
          & Acc.$\downarrow$ & Comp.$\downarrow$ & Time$\downarrow$
          & Acc.$\downarrow$ & Comp.$\downarrow$ & Time$\downarrow$
          & Acc.$\downarrow$ & Comp.$\downarrow$ & Time$\downarrow$ \\
\hline
VGGT*~\cite{wang2025vggt}       
& 1.00 & 1.00 
& 0.010 & 0.009 & 135.1s 
& 0.011 & 0.011 & 177.5s
& 0.028 & 0.022 & 724.6s \\
FastVGGT~\cite{shen2025fastvggt}
& 0.34 & 0.34 
& 0.014 & 0.020 & 51.2s 
& 0.012 & 0.011 & 52.3s
& 0.027 & 0.021 & 175.2s \\ 
\rowcolor{green!15}
VGGT*+HTTM             
& \textbf{0.20} & 0.30 
& 0.010 & 0.008 & 26.4s 
& 0.011 & 0.010 & 35.8s
& 0.027 & 0.021 & \textbf{102.8s} \\
\hline
\end{tabular}
\caption{3D reconstruction performance with longer sequence input. With longer sequence inputs, HTTM constantly shows similar performance to the original VGGT with substantially shorter latency.}
\label{tab:longer}
\end{table*}

\noindent \textit{(ii) Cost increases linearly with block size.} 

Let $n_s^{(k)}=|\mathcal{S}_k|$ and $n_d^{(k)}=|\mathcal{D}_k|$, with $n_s^{(k)}+n_d^{(k)}=n_b$.  
Computing similarities inside block $k$ costs $\Theta(n_s^{(k)} n_d^{(k)} d)$ operations, where $d$ is the head dimension.

\noindent Assuming a fixed source/destination ratio:
\[
n_s^{(k)}=\alpha n_b\,\quad n_d^{(k)}=(1-\alpha)n_b\,,
\]
We have that
\[
n_s^{(k)} n_d^{(k)} = \alpha(1-\alpha) n_b^2\,.
\]
With $K\approx N/n_b$ blocks, the total cost can be approximated as
\[
\sum_{k=1}^K \Theta(\alpha(1-\alpha)n_b^2 d)
\approx  \Theta\big(\alpha(1-\alpha)\frac{N}{n_b}\,n_b^2 d\big)
= O(N n_b d),
\]
which grows linearly with block size $n_b$.

Combining both parts, larger blocks always improve (or maintain) merging quality but incur proportionally higher computational costs.

\subsection{Experiments on Longer Sequence}
In this section, we show more experiments on longer sequences with NRGBD~\cite{azinovic2022neural} and ScanNet~\cite{dai2017scannet} dataset in Table~\ref{tab:longer}.

\paragraph{Setup}
For the NRGBD dataset, we evaluate HTTM by sampling keyframes every 3 frames. For the ScanNet dataset, we randomly select 15 scenes with over 2000 input frames and sample keyframes every 2 frames. The setup of FastVGGT and our HTTM is the same as in Sec.~\ref{sec:3d_recon}.

\paragraph{Results}
As shown in Table~\ref{tab:longer}, with longer sequence inputs, HTTM constantly shows similar performance to the original VGGT with substantially shorter latency.

\begin{figure*}[h]
    \centering
    \begin{subfigure}{0.49\linewidth}
        \centering
        \includegraphics[width=\linewidth]{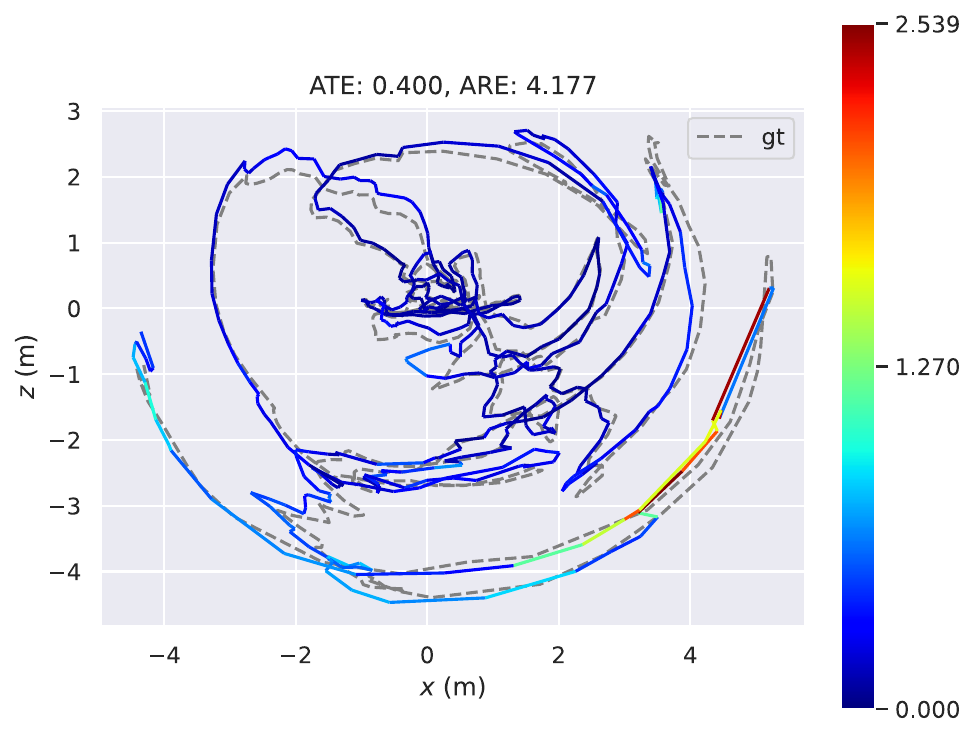}
        \caption{FastVGGT}
        \label{fig:error_fastvggt}
    \end{subfigure}
    \begin{subfigure}{0.49\linewidth}
        \centering
        \includegraphics[width=\linewidth]{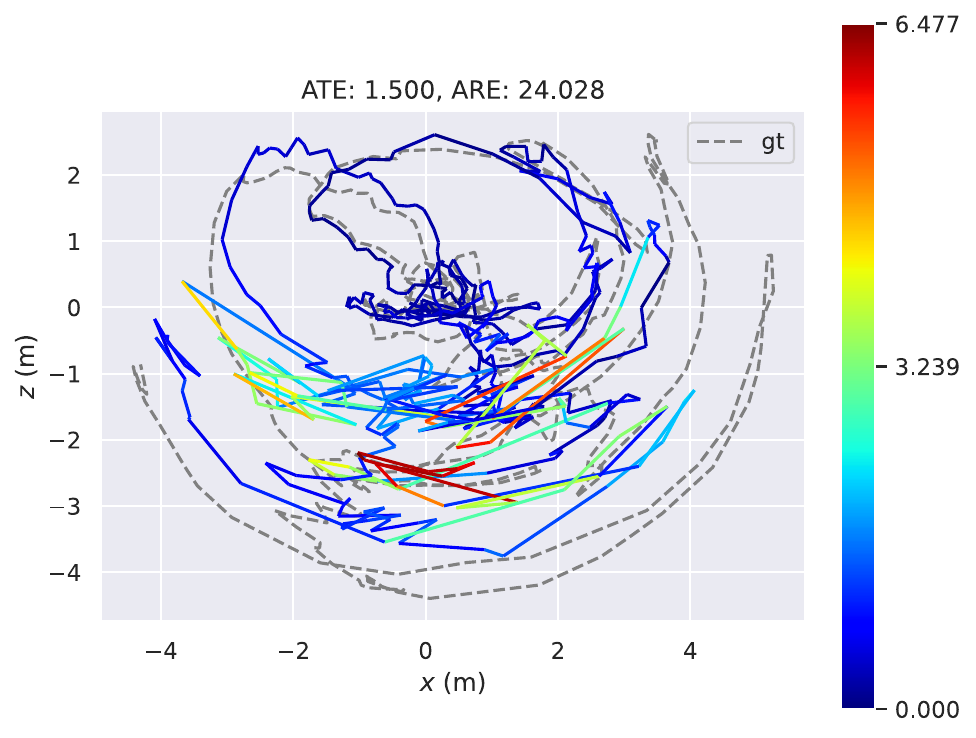}
        \caption{VGGT}
        \label{fig:error_vggt}
    \end{subfigure}
    
    \begin{subfigure}{0.49\linewidth}
        \centering
        \includegraphics[width=\linewidth]{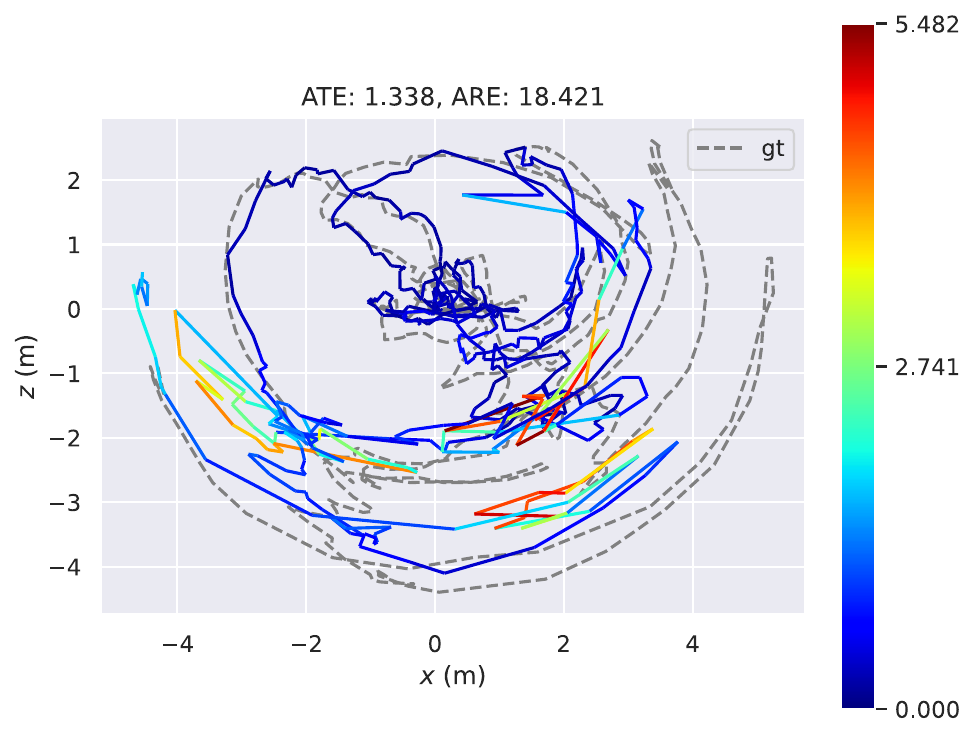}
        \caption{HTTM without first frame anchoring}
        \label{fig:error_no}
    \end{subfigure}
    \begin{subfigure}{0.49\linewidth}
        \centering
        \includegraphics[width=\linewidth]{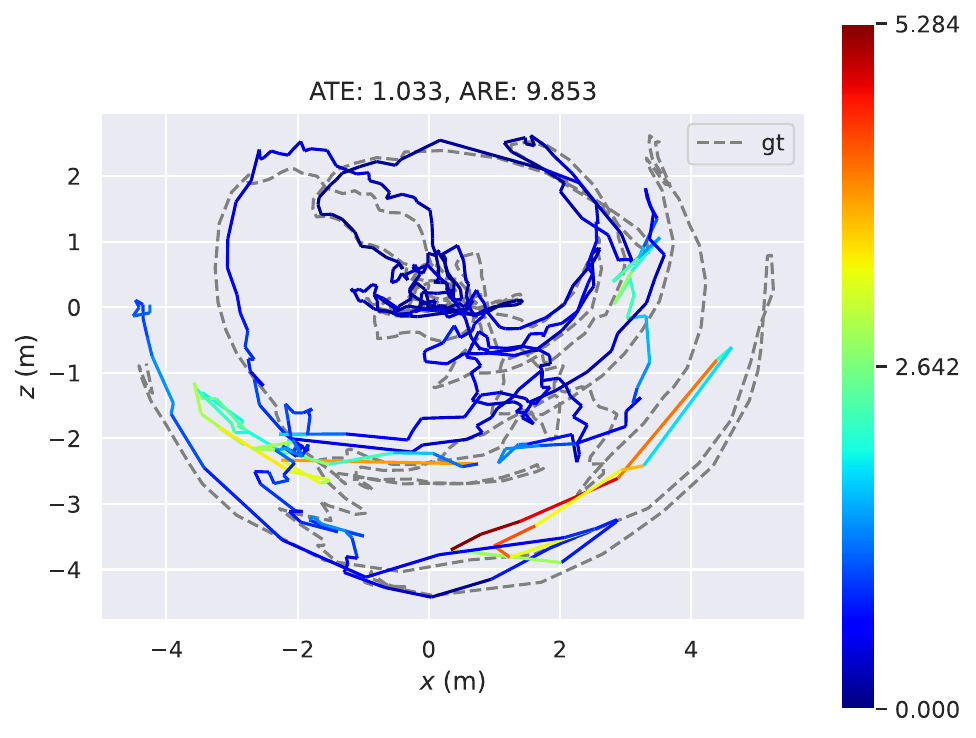}
        \caption{HTTM$^{\star}$ with 10 frames of temporal merging}
        \label{fig:error_10}
    \end{subfigure}
    
    \begin{subfigure}{0.49\linewidth}
        \centering
        \includegraphics[width=\linewidth]{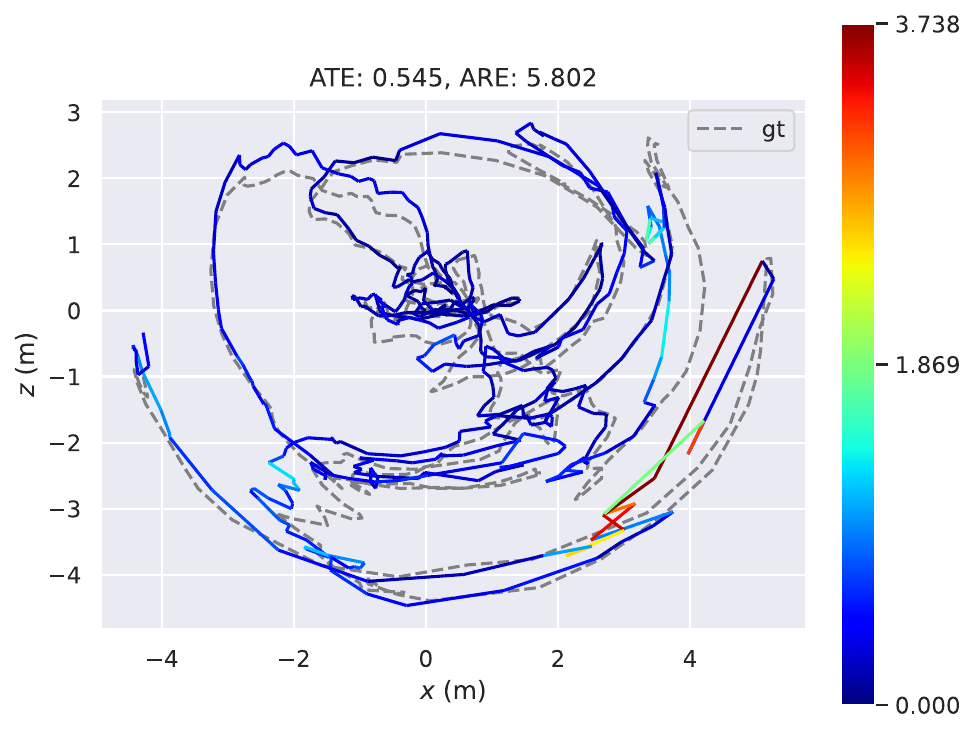}
        \caption{HTTM$^{\star}$ with 30 frames of temporal merging}
        \label{fig:error_30}
    \end{subfigure}
    \begin{subfigure}{0.49\linewidth}
        \centering
        \includegraphics[width=\linewidth]{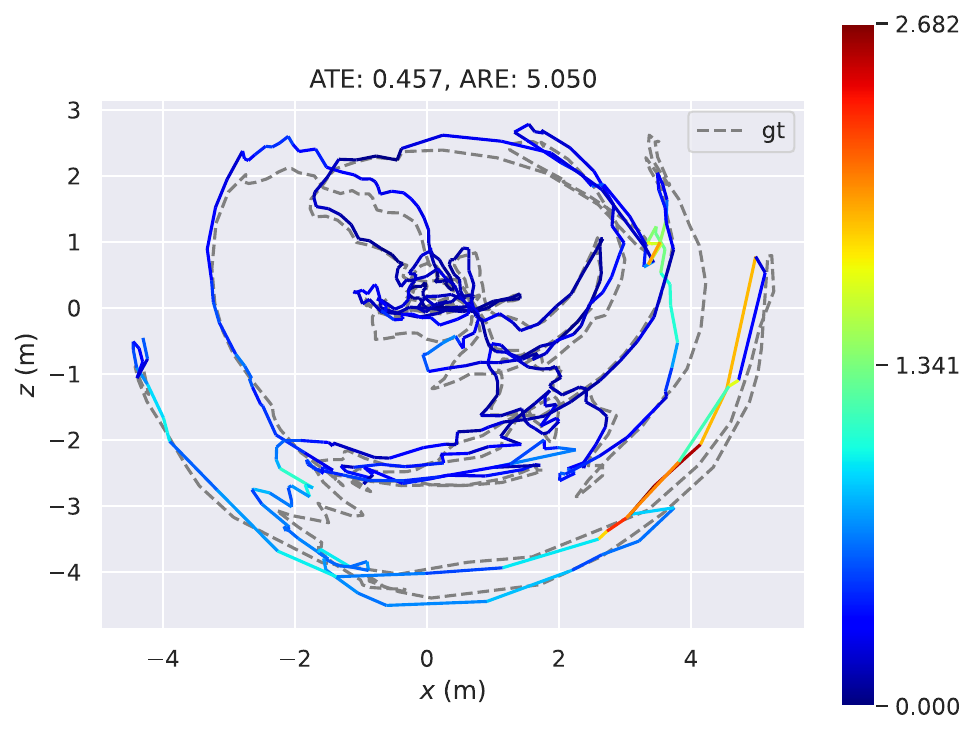}
        \caption{HTTM$^{\star}$ with 40 frames of temporal merging}
        \label{fig:error_50}
    \end{subfigure}

    \caption{Comparison of camera pose estimation performance between FastVGGT, VGGT, HTTM without first frame anchoring, and HTTM with first frame anchoring (denoted as HTTM$^{\star}$) across different numbers of temporal frames. Colors indicate the deviation from the ground-truth camera trajectory.}
    \label{fig:error}
\end{figure*}

\subsection{Error Mitigation}
In this section, we discuss the error mitigation effect of token merging on VGGT reported by FastVGGT~\cite{shen2025fastvggt}. 

The original VGGT shows a large error in camera pose estimation when the camera movement is high. For example, in Fig.~\ref{fig:error}, we visualize the camera pose estimation error on a big scene from ScanNet with large camera movement, using 500 keyframes sampled every 10 frames. Compared to FastVGGT (Fig.~\ref{fig:error_fastvggt}), the original VGGT (Fig.~\ref{fig:error_vggt}) shows much higher error in camera pose estimation. HTTM shows a smaller error compared to the original VGGT, but still higher than FastVGGT. Although FastVGGT offers a discussion of the observed improvement, the specific mechanism responsible for the enhanced error mitigation ability remains insufficiently clarified.

In order to understand this higher error mitigation ability of FastVGGT, we further investigated it, and we found that the error mitigation effect comes from the first-frame anchoring design in FastVGGT. FastVGGT assigns all tokens in the first frame as \texttt{dst} tokens, referring to them as “Reference Tokens” to preserve their strong representativeness. However, we find that the crucial factor is to reduce tokens from subsequent frames that are highly similar to those in the first frame. As shown in Fig.~\ref{fig:error_10},\ref{fig:error_30},\ref{fig:error_50}, by adding first frame anchoring and allowing more temporal merging (so that more tokens from subsequent frames can be merged to the first frame), HTTM achieves a similar error mitigation effect to FastVGGT. 
We speculate that tokens from later frames can mislead the Global Attention module. Because these tokens are highly similar to first-frame tokens, the model may incorrectly treat them as part of the reference frame, weakening the coordinate anchor and amplifying drift. By explicitly designating all first-frame tokens as \texttt{dst} tokens and merging highly similar tokens from subsequent frames into them, the ambiguity is suppressed and the reference frame remains stable.

Note that in a large scene with long-sequence input, the tokens from the first frames consist less than 1\% of the whole token set, so activating first-frame anchoring introduces negligible overhead for HTTM.

\newpage
\section{Algorithm Pseudo Codes}
In this section, we provide the pseudo code of some algorithms in the hope of helping people who are interested in reimplementing our work.
\subsection{HTTM Accelerated Global Attention}

\begin{algorithm}[h]
\caption{\textsc{HTTM}: Global Attention with Temporal Reordering and Blockwise Token Merging}
\label{alg:httm}
\begin{algorithmic}[1]
\Require Input tokens $\bm{X} \in \mathbb{R}^{L \times D}$, 
query cluster number $k_q$, key cluster number $k_k$,
spatial block size $B_s$, temporal block size $B_t$,
query outlier quantile $\gamma_q$, key outlier quantile $\gamma_k$
\Ensure Output tokens $\bm{O} \in \mathbb{R}^{L \times D}$

\State Compute multi-head projections 
$\bm{Q}, \bm{K}, \bm{V} \in \mathbb{R}^{H \times L \times E}$ from $\bm{X}$
\State $B \leftarrow B_s \cdot B_t$
\State Compute temporal reordering index $\pi \leftarrow \textsc{TemporalReorder}(L, B_s, B_t)$
\State $\tilde{\bm{Q}} \leftarrow \textsc{Reorder}(\bm{Q}, \pi)$,
       $\tilde{\bm{K}} \leftarrow \textsc{Reorder}(\bm{K}, \pi)$,
       $\tilde{\bm{V}} \leftarrow \textsc{Reorder}(\bm{V}, \pi)$
\State $\bm{C_q} \leftarrow \textsc{BlockwiseToMe}(\tilde{\bm{Q}}, B, k_q)$
\State $\bm{C_k} \leftarrow \textsc{BlockwiseToMe}(\tilde{\bm{K}}, B, k_k)$
\State $\bm{Q}_m, \bm{M_q} \leftarrow$
\State \hspace{1em} $\textsc{BlockAggregateQuantile}(\tilde{\bm{Q}}, \bm{C_q}, B, k_q, \gamma_q)$
\State $\bm{K}_m \leftarrow \textsc{BlockAggregate}(\tilde{\bm{K}}, \bm{C_k}, B, k_k)$
\State $\bm{V}_m \leftarrow \textsc{BlockAggregate}(\tilde{\bm{V}}, \bm{C_k}, B, k_k)$
\State $\bm{O}_m \leftarrow \textsc{Attention}(\bm{Q}_m, \bm{K}_m, \bm{V}_m)$
\State $\tilde{\bm{O}} \leftarrow \textsc{UnmergeByCluster}(\bm{O}_m, \bm{C_q}, \bm{M_q})$
\State $\tilde{\bm{O}}[\bm{M_q}] \leftarrow \textsc{Attention}(\tilde{\bm{Q}}[\bm{M_q}], \bm{K}_m, \bm{V}_m)$
\State $\bm{O} \leftarrow \textsc{InverseReorder}(\tilde{\bm{O}}, \pi)$
\State $\bm{O} \leftarrow \textsc{OutputProj}(\bm{O})$
\State \Return $\bm{O}$
\end{algorithmic}
\end{algorithm}

\vspace{20em}

\subsection{Adaptive Outlier Filtering}

\begin{algorithm}[h]
\caption{\textsc{BlockAggregateQuantile}: Blockwise Token Aggregation with Quantile-based Outlier Filtering}
\label{alg:agg}
\begin{algorithmic}[1]
\Require Tokens $\bm{Q} \in \mathbb{R}^{H \times L \times E}$, 
cluster assignments $\bm{C_q} \in \mathbb{Z}^{H \times L}$, 
block size $B$, cluster number $k$, quantile $\gamma$
\Ensure Aggregated tokens $\bm{Q_c}$, outlier mask $\bm{M} \in \{True, False\}^{H \times L}$

\State Partition $\bm{Q}$ and $\bm{C_q}$ into $H \times \lceil L/B \rceil$ blocks
\State Initialize centroid tensor $\bm{Q_c}$, error tensor $\bm{\varepsilon} \in \mathbb{R}^{H \times L}$, and outlier mask $\bm{M} \leftarrow \text{False}$

\For{each block $i$}
    \State Compute cluster centroids $\bm{Q_c}^{(i)}$ from $\bm{Q}^{(i)}$ using $\bm{C_q}^{(i)}$
    \For{each token $j$ in block $i$}
        \State $c \leftarrow \bm{C_q}^{(i)}[j]$
        \If{$c \ge 0$}
            \State $\bm{\varepsilon}^{(i)}[j] \leftarrow \|\bm{Q}^{(i)}[j] - \bm{Q_c}^{(i)}[c]\|_2$
        \Else
            \State $\bm{\varepsilon}^{(i)}[j] \leftarrow 0$
        \EndIf
    \EndFor
\EndFor

\State $\tau \leftarrow \textsc{Quantile}(\bm{\varepsilon}, \gamma)$

\For{each block $i$}
    \State Mark tokens with $\bm{\varepsilon}^{(i)}[j] > \tau$ as outliers:
    $\bm{C_q}^{(i)}[j] \leftarrow -1,\ \bm{M}^{(i)}[j] \leftarrow \text{True}$
    \State Recompute only the centroids of clusters affected by outlier removal
    \State Append final centroids and retained outliers of block $i$ to $\bm{Q_c}$
\EndFor

\State \Return $\bm{Q_c}, \bm{M}$
\end{algorithmic}
\end{algorithm}
\end{document}